\documentclass{article}

\usepackage{iclr2025_conference,times}

\usepackage{amsmath,amsfonts,bm}

\def\eqref#1{equation~\ref{#1}}

\def\1{\bm{1}}

\DeclareMathAlphabet{\mathsfit}{\encodingdefault}{\sfdefault}{m}{sl}
\SetMathAlphabet{\mathsfit}{bold}{\encodingdefault}{\sfdefault}{bx}{n}

\usepackage{amsmath}
\usepackage{amssymb}
\usepackage{bbm}

\usepackage{booktabs}
\usepackage{tabularx}
\usepackage{multirow}
\usepackage{adjustbox}
\usepackage{graphicx}
\usepackage{subcaption}
\usepackage{float}

\usepackage[table]{xcolor}
\usepackage[most]{tcolorbox}
\usepackage{ragged2e}

\usepackage{algorithm}
\usepackage{algpseudocode}

\usepackage{pifont}
\usepackage{fontawesome5}
\usepackage{textcomp}
\usepackage{textgreek}

\usepackage{url}
\usepackage{hyperref}

\providecommand{\tblYes}{\ding{51}}
\providecommand{\tblNo}{\ding{55}}

\providecommand{\faListAlt}{\faClipboardList}

\definecolor{corpusgray}{HTML}{EEF1F4}
\definecolor{simulationorange}{HTML}{F8E8DA}
\definecolor{interactiveblue}{HTML}{E8F1F8}
\definecolor{triagegreen}{HTML}{E7F2E8}
\definecolor{oursgreen}{HTML}{CFE8D2}

\definecolor{physicianblue}{RGB}{232,242,252}
\definecolor{physicianborder}{RGB}{126,174,213}
\definecolor{patientgreen}{RGB}{234,247,237}
\definecolor{patientborder}{RGB}{126,188,137}

\definecolor{advcol}{HTML}{C2E0C6}
\definecolor{profcol}{HTML}{FFF4CC}
\definecolor{devcol}{HTML}{FFD6D6}
\definecolor{inccol}{gray}{0.9}

\title{
ELICITED: EHR-grounded Longitudinal Interactive Conversations for Information-seeking Triage Evaluation and Decision-making
}

\author{
Haohao Zhu\footnotemark[2]\\
University of Michigan\\
\href{mailto:haohaoz@umich.edu}{\texttt{haohaoz@umich.edu}}
\And
Xiaolin Shi\\
Ellipsis Health\\
\href{mailto:xiaolin.shi@ellipsishealth.com}{\texttt{xiaolin.shi@ellipsishealth.com}}
\And
Jiayu Zhou\\
University of Michigan\\
\href{mailto:jiayuz@umich.edu}{\texttt{jiayuz@umich.edu}}
}
\iclrfinalcopy

\begin{document}

\maketitle

\begingroup
\renewcommand{\thefootnote}{\fnsymbol{footnote}}
\footnotetext[2]{Work done during an internship at Ellipsis Health.}
\endgroup

\fancyhead{}


\begin{abstract}
Emergency-department (ED) triage requires clinicians to rapidly identify
patients who need immediate attention, determine who can safely wait, and
prioritize limited clinical resources. At presentation, however, the available
information is often incomplete and may be limited to a brief chief complaint
and initial vital signs. Clinically important details—including symptom onset
and progression, associated symptoms, relevant medical history, and medication
use—are frequently obtained through focused conversation. Effective triage
therefore involves not only assigning an acuity level, but also identifying
information gaps, asking appropriate follow-up questions, and updating the
assessment as new evidence becomes available. Clear communication further
supports patient understanding of the immediate plan and facilitates timely
reporting of clinical deterioration while waiting. Most existing ED benchmarks evaluate acuity prediction from a fixed clinical
snapshot. This formulation is well suited to measuring predictive performance
once patient information has been assembled, but it does not fully capture the
interactive process through which triage-relevant evidence is elicited and
interpreted. Medical dialogue datasets offer complementary opportunities for
studying clinical communication, although dialogue statements are not always
linked to temporally ordered events in the corresponding electronic health
record (EHR). Event-level and temporal grounding are therefore important for
evaluating whether a conversational system obtains clinically relevant
information, uses supported evidence, and respects what could reasonably have
been known at the time of triage. We introduce \textbf{EHR2Dial-Triage}, an agentic conversation-generation
framework and benchmark grounded in MIMIC-IV-ED. The framework constructs
triage conversations under explicit role-based and temporal information
boundaries. Each accepted patient disclosure is linked to its supporting EHR
event and to the first dialogue turn at which it becomes available.
EHR2Dial-Triage supports controlled evaluation of information elicitation,
evidence use, five-level Emergency Severity Index prediction, and patient
communication across models, and patient personas.
Together, these capabilities provide a structured setting for studying
conversational triage as a dynamic process of clinical information acquisition,
reasoning, and communication.

\textbf{Code is available at} \url{https://github.com/illidanlab/elicited}.
\end{abstract}

\section{Introduction}
\label{sec:introduction}

Emergency department (ED) triage is the first stage at which limited clinical
resources must be allocated according to patient need. Its purpose is not to
establish a definitive diagnosis, but to determine who requires immediate
evaluation, who can safely wait, and what level of care and resource use may be
needed. These decisions affect both patient safety and ED operations.
Under-triage may delay treatment for time-sensitive conditions, whereas
over-triage may direct staff attention, treatment spaces, monitoring, and
diagnostic resources away from other patients. The Emergency Severity Index
(ESI) reflects this dual objective by incorporating both clinical urgency and
anticipated resource use \citep{wuerz2000esi,tanabe2004esi}. In a retrospective
study of more than five million ED encounters, only 32.2\% met the study's
operational definition of ESI mistriage, including 3.3\% classified as
under-triage \citep{sax2023mistriage}. Accurate triage is therefore important
not only for identifying high-risk patients, but also for supporting efficient
and equitable resource allocation across the ED.

The difficulty is that triage decisions must be made quickly and from incomplete information. At arrival, clinicians may have only a brief chief complaint, initial vital signs, and immediately observable features of the patient's condition. The urgency of the same presenting complaint may differ substantially depending on symptom onset, progression, severity, associated symptoms, relevant medical history, recent treatment, or medication use \citep{ena2023esi,reay2024triage}. Much of this information is obtained through the triage interview \citep{johnson2018interruptions}. The clinician must therefore identify which information is missing, ask questions whose answers may change the urgency assessment, and update the decision as new evidence becomes available. Because triage time is limited, the objective is not to collect a complete medical history, but to acquire the information most relevant to prioritization and resource use.

Effective triage also depends on clear patient communication. Patients must be able to describe their symptoms, understand the immediate plan, and recognize when worsening symptoms should be reported while they wait. Clear and focused communication can support this information exchange while helping patients understand what will happen next. Communication, information provision, waiting time, and interactions with triage staff are consistently identified as important determinants of the ED and triage experience \citep{sonis2018patientexperience,janerka2024triageexperience}. Effective triage therefore involves both focused information acquisition and appropriate patient-facing communication.

Recent progress in LLM-based clinical agents makes this interactive setting increasingly feasible. Models can sustain multi-turn clinical conversations, ask follow-up questions, incorporate newly disclosed information, and reason over evolving clinical context \citep{li2024mediq,tu2025amie,schmidgall2026agentclinic,ferber2026mira}. This creates an opportunity to study triage not only as prediction from preassembled clinical inputs, but also as an interactive process in which the model must decide what information to seek before making an acuity assessment.

Conversational triage differs from an extended diagnostic consultation. Its immediate objective is to assess acuity and prioritize patients for emergency care rather than to establish a definitive diagnosis \citep{ena2023esi,gorick2023triagedecision}. Triage assessments are necessarily focused and time-constrained, with the goal of assigning patients to an appropriate level of care according to clinical urgency \citep{zou2026triageefficiency}. A conversational agent should therefore identify which questions are most relevant to the urgency assessment rather than attempt to reconstruct the entire clinical case. This motivates evaluating not only the final acuity prediction, but also what information the agent elicits, how that information affects the assessment, and how clearly the resulting decision is communicated to the patient.

EHR data provide a natural basis for constructing patient simulations because they preserve case-specific symptoms, medical history, medications, measurements, and the temporal progression of real encounters. Prior work has shown that conditioning synthetic clinical dialogue on patient records can improve factuality, while EHR-derived patient simulators use case-specific profiles to support and evaluate factual consistency \citep{das2024syndial,kyung2025patientsim}. Grounding is particularly important in clinical dialogue, where fluent generation may otherwise introduce plausible but unsupported information \citep{asgari2025hallucination,wu2025medkp}. We therefore use the EHR not only to construct patient profiles, but also to constrain what information can enter the conversation. Each accepted disclosure must be supported by an eligible EHR event and be valid at the current point in the encounter, limiting unsupported or hallucinated content while making factual and temporal errors directly auditable.

We introduce \textbf{EHR2Dial-Triage}, an EHR-to-dialogue framework and
benchmark for studying multi-turn ED triage under explicit role-based and
temporal information boundaries. Each MIMIC-IV-ED encounter
\citep{johnson2023mimicived} is divided into information available at the
beginning of triage, patient-reportable information that may be elicited during
the conversation, and future or evaluation-only information that remains
hidden. A clinician model asks focused questions using the currently available
evidence and dialogue history, while a patient model responds using eligible
case information. A verifier checks each proposed disclosure and records its
supporting EHR event and first accepted dialogue turn.

We use EHR2Dial-Triage to evaluate two complementary LLM capabilities in conversational triage. First, with the patient simulator held fixed, we examine what questions different LLMs ask as triage clinicians, what source-supported information they elicit, and how the accumulated information changes an independent ESI reader's assessment. Second, given the same completed dialogue and the information disclosed during the interaction, we evaluate whether LLMs can predict the recorded five-level ESI and explain the rationale for that triage decision to the patient. These settings separate information acquisition from the subsequent use and communication of the acquired evidence. Across models, performance on these capabilities does not consistently align, suggesting that conversational triage should be evaluated beyond final acuity prediction alone.

\begin{table*}[t]
\centering
\scriptsize
\setlength{\tabcolsep}{2.6pt}
\renewcommand{\arraystretch}{1.12}

\caption{
\textbf{Comparison with representative clinical dialogue and interaction
resources.}
\textit{Cases} denotes source cases, patient profiles, or evaluation scenarios;
\textit{On Demand} indicates dialogues generated dynamically rather than
released as a fixed corpus. A reusable generator requires publicly released,
runnable code for generating complete multi-turn dialogues from new cases.
EHR grounding requires patient-level EHR records; event--turn lineage requires
a machine-readable link from each disclosed fact to its source event and first
accepted dialogue turn; and temporal tracing requires explicit separation of
initially visible, patient-elicitable, and future information.
\textsc{Partial} denotes mixed EHR- and non-EHR-grounded components.
}
\label{tab:benchmark_comparison}

\begin{adjustbox}{max width=\textwidth}
\begin{tabular}{@{}lllrrcccc@{}}
\toprule

Paper Title
& Source
& Downstream Task
& \shortstack{Clinical\\Cases}
& \shortstack{Released\\Dialogue}
& \shortstack{Reusable\\Generator}
& \shortstack{EHR\\Grounding}
& \shortstack{Event--turn\\Lineage}
& \shortstack{Temporal\\Tracing} \\

\midrule

\rowcolor{corpusgray}
\multicolumn{9}{l}{\textit{Clinical dialogue corpora}} \\

MedDialog~\citep{zeng2020meddialog}
& Online consultations
& Response generation
& -- & 3.66M
& \tblNo & \tblNo & \tblNo & \tblNo \\

IMCS-21~\citep{chen2023imcs21}
& Pediatric consultations
& Dialogue modeling
& -- & 4,116
& \tblNo & \tblNo & \tblNo & \tblNo \\

ReMeDi~\citep{yan2022remedi}
& Online consultations
& Dialogue modeling
& -- & 96,965
& \tblNo & \tblNo & \tblNo & \tblNo \\

MediTOD~\citep{saley2024meditod}
& Staged clinical interviews
& Dialogue modeling
& -- & 213
& \tblNo & \tblNo & \tblNo & \tblNo \\

\addlinespace[2pt]
\rowcolor{simulationorange}
\multicolumn{9}{l}{
\textit{Document-conditioned dialogue and patient simulation}
} \\

NoteChat~\citep{wang2024notechat}
& Case reports and dialogues
& Dialogue generation
& 10,000 & 10,000
& \tblYes & \tblNo & \tblNo & \tblNo \\

Note2Chat~\citep{zhou2026note2chat}
& MIMIC-IV-Note
& Dialogue generation
& 4,972 & 8,944
& \tblYes & \tblYes & \tblNo & \tblNo \\

PatientSim~\citep{kyung2025patientsim}
& MIMIC-IV
& Patient simulation
& 170 & \textit{On Demand}
& \tblYes & \tblYes & \tblNo & \tblNo \\

\addlinespace[2pt]
\rowcolor{interactiveblue}
\multicolumn{9}{l}{\textit{Interactive clinical benchmarks}} \\

MediQ~\citep{li2024mediq}
& Medical QA and clinical cases
& Interactive diagnosis
& 12,863 & \textit{On Demand}
& \tblYes & \tblNo & \tblNo & \tblNo \\

APP~\citep{zhu2025app}
& ReMeDi-derived cases
& Interactive diagnosis
& 100 & --
& \tblNo & \tblNo & \tblNo & \tblNo \\

EPAG~\citep{seo2026epag}
& Synthetic clinical profiles
& Interactive diagnosis
& 520 & \textit{On Demand}
& \tblYes & \tblNo & \tblNo & \tblNo \\

MINT~\citep{fang2026mint}
& Medical QA cases
& Interactive diagnosis
& 1,035 & --
& \tblNo & \tblNo & \tblNo & \tblNo \\

AMIE~\citep{tu2025amie}
& OSCE-style scenarios
& Interactive diagnosis
& 159 & --
& \tblNo & \tblNo & \tblNo & \tblNo \\

MeDxBench~\citep{sanghvi2026medxagent}
& Public diagnostic datasets
& Interactive diagnosis
& 4,421 & --
& \tblNo & \tblNo & \tblNo & \tblNo \\

AgentClinic~\citep{schmidgall2026agentclinic}
& Mixed clinical benchmarks
& Clinical-agent evaluation
& Suite & \textit{On Demand}
& \tblYes & \textsc{Partial} & \tblNo & \tblNo \\

3MDBench~\citep{sviridov2025three}
& Telemedicine cases
& Multimodal diagnosis
& 2,996 & \textit{On Demand}
& \tblYes & \tblNo & \tblNo & \tblNo \\

MedAgentBench~\citep{jiang2025medagentbench}
& Deidentified STARR EHR
& EHR-agent evaluation
& 100 & --
& \tblNo & \tblYes & \tblNo & \tblNo \\

MIRA~\citep{ferber2026mira}
& MIMIC-IV
& Interactive diagnosis
& 574 & --
& \tblNo & \tblYes & \tblNo & \tblNo \\

\addlinespace[2pt]
\rowcolor{triagegreen}
\multicolumn{9}{l}{\textit{Conversational triage benchmarks}} \\

Conv.\ Triage~\citep{rashidian2025conversational}
& Longitudinal EHR
& Conversational triage
& 519 & --
& \tblNo & \tblYes & \tblNo & \tblNo \\

TriageSim~\citep{srirag2026triagesim}
& Mixed triage sources
& Conversational triage
& 814 & On Demand
& \tblYes & \textsc{Partial} & \tblNo & \tblNo \\

\midrule

\rowcolor{oursgreen}
\textbf{EHR2Dial-Triage}
& \textbf{MIMIC-IV-ED}
& \textbf{Conversational triage}
& \textbf{4,041}
& On Demand
& \tblYes & \tblYes & \tblYes & \tblYes \\

\bottomrule
\bottomrule
\end{tabular}
\end{adjustbox}
\end{table*}

\section{Related Work}
\label{sec:related-work}

EHR2Dial-Triage draws on four connected areas of research: ED acuity
prediction, clinical dialogue generation, interactive clinical information
acquisition, and conversational triage. These areas address different stages
of the broader problem, ranging from estimating urgency from assembled records
to constructing patient agents and evaluating multi-turn clinical
interactions. Table~\ref{tab:benchmark_comparison} summarizes representative
resources and highlights the provenance and timing mechanisms that are
particularly relevant to the present benchmark.

\subsection{Emergency-Department Triage and Acuity Prediction}

Computational ED triage has commonly been studied as supervised inference from
structured records, clinical text, or combinations of both. Using
MIMIC-IV-ED \citep{johnson2023mimicived},
\citet{xie2022edbenchmark} standardized prediction tasks for hospitalization,
critical outcomes, and 72-hour ED reattendance. KATE predicts the five-level
Emergency Severity Index \citep{wuerz2000esi} using structured and textual EHR
features \citep{ivanov2021kate}. More recent studies have examined large
language models for pairwise acuity comparison from ED notes
\citep{williams2024llmtriage} and for direct ESI prediction from
MIMIC-IV-ED records \citep{lafuente2025automatedtriage}.

This literature establishes clinically meaningful targets, evaluation
protocols, and strong predictive baselines for ED risk stratification. It also
provides the downstream decision task used in our benchmark. EHR2Dial-Triage
extends this setting by treating the record presented to the predictor as the
result of an information-acquisition process: the model first interacts with a
patient agent and then evaluates urgency using the evidence accumulated during
that interaction.

\subsection{Clinical Dialogue Data and EHR-Grounded Simulation}

Clinical dialogue resources span naturally collected consultations, staged
encounters, and conversations synthesized from clinical documents. MedDialog
provides large-scale online consultations for response generation
\citep{zeng2020meddialog}. IMCS-21, ReMeDi, and MediTOD support research on
dialogue understanding, dialogue policy, symptom inquiry, and clinical
language generation
\citep{chen2023imcs21,yan2022remedi,saley2024meditod}. These datasets have
enabled the development and evaluation of models that represent clinical
dialogue structure and generate medically relevant responses.

Document-conditioned approaches connect dialogue generation more directly to
case-specific clinical information. NoteChat synthesizes patient--physician
conversations from clinical case reports \citep{wang2024notechat}, while
Note2Chat converts MIMIC-IV notes into multi-turn history-taking dialogues for
diagnostic reasoning \citep{zhou2026note2chat}. PatientSim constructs
persona-conditioned patient agents from profiles derived from MIMIC-IV,
MIMIC-IV-ED, and MIMIC-IV-Note \citep{kyung2025patientsim}. These approaches
demonstrate how clinical records can support coherent, case-conditioned patient
simulation.

EHR2Dial-Triage builds on this record-conditioned paradigm but adopts a finer
unit of provenance for benchmark analysis. In addition to conditioning a
dialogue on an encounter, the framework records the source event supporting
each accepted disclosure and the first turn at which the disclosure becomes
available. This representation is designed for evaluating the sequence through
which information is acquired, rather than only the coherence of the completed
conversation.

\subsection{Interactive Clinical Information Acquisition}

Interactive clinical benchmarks study how models gather additional evidence
before producing a clinical assessment. DDXPlus supports symptom inquiry and
differential diagnosis using large-scale synthetic patient profiles
\citep{fansitchango2022ddxplus}. CRAFT-MD evaluates conversational history
taking across multiple specialties \citep{johri2025craftmd}. MediQ and Ask
Patients with Patience examine follow-up question selection and the relationship
between information seeking and diagnostic performance
\citep{li2024mediq,zhu2025app}. EPAG evaluates collection of the history of
present illness with reference to diagnostic guidelines
\citep{seo2026epag}. MINT incrementally reveals evidence to study premature
commitment and self-correction \citep{fang2026mint}, while MeDxBench evaluates
open-ended diagnostic interviews with case-grounded patient agents
\citep{sanghvi2026medxagent}.

Other environments broaden interactive evaluation to communication,
multimodal inputs, and tool-mediated clinical workflows. AMIE evaluates
diagnostic dialogue and patient communication \citep{tu2025amie}.
AgentClinic provides an environment for clinical-agent evaluation across
multiple tasks \citep{schmidgall2026agentclinic}, and 3MDBench incorporates
multimodal telemedicine interactions \citep{sviridov2025three}.
MedAgentBench evaluates retrieval and actions in a FHIR-compatible EHR
environment \citep{jiang2025medagentbench}, while MIRA combines an
EHR-grounded patient agent with diagnostic and management decisions
\citep{ferber2026mira}.

These benchmarks establish information seeking as a measurable clinical-agent
capability and provide several complementary evaluation paradigms.
EHR2Dial-Triage focuses specifically on the relationship among question
selection, newly elicited evidence, and urgency assessment. Its temporal
partitioning additionally distinguishes evidence that may be elicited from the
patient during triage from events that become available only later in the ED
encounter.

\subsection{Conversational Triage and Benchmark Positioning}

The work most directly related to EHR2Dial-Triage applies patient simulation to
triage assessment. \citet{rashidian2025conversational} evaluate multi-turn
interactions between an EHR-derived patient simulator and a symptom-checking
agent, with attention to response consistency and clinical-summary quality.
TriageSim generates persona-conditioned text and audio conversations from
structured ED records and triage teaching cases, and evaluates acuity
classification across text, automatic-speech-recognition, and audio inputs
\citep{srirag2026triagesim}. These studies establish conversational triage as
an important evaluation setting and demonstrate the value of realistic patient
simulation, persona variation, and multimodal interaction.

EHR2Dial-Triage studies a complementary aspect of conversational triage:
auditable information acquisition from temporally organized patient-level EHR
events. The benchmark represents what is visible at the beginning of triage,
what may be elicited from the patient, and what remains unavailable until later
in the encounter. It also records event--turn lineage for each accepted
disclosure. These design choices make it possible to analyze not only the final
acuity decision, but also which evidence was acquired, how it entered the
conversation, and whether it was available at the appropriate stage of the
encounter. A case-level illustration of these differences across representative resource
designs is provided in Appendix Figure~\ref{fig:resource-design-comparison}.

\begin{figure*}[t]
    \centering
    \includegraphics[width=\textwidth]{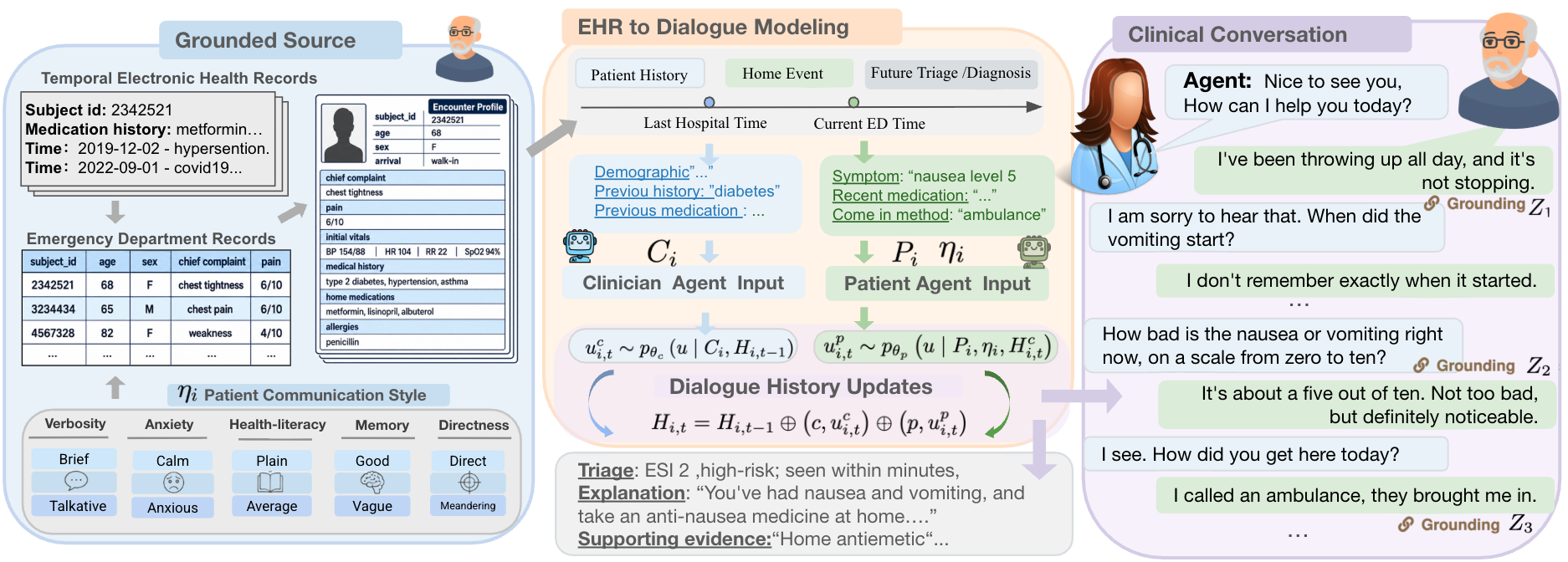}
    \caption{\textbf{Overview of EHR2Dial-Triage.}
    Each ED encounter is represented by clinician-visible information
    \(C_i\), patient-reportable information \(P_i\), and a non-clinical
    patient persona \(\eta_i\). Role-specific clinician and patient models
    generate an alternating triage dialogue under asymmetric information
    access. Each accepted disclosure remains linked to its supporting EHR
    source and first accepted speaker turn, enabling exact tracing of
    information acquisition throughout the dialogue.}
    \label{fig:ehr2dial-workflow}
\end{figure*}

\section{EHR-to-Dialogue Triage Modeling}
\label{sec:method}

\subsection{Information Partition}
\label{sec:partition}

We formulate EHR-to-dialogue generation as a sequential interaction between
two role-specific language models with asymmetric information access. For ED
encounter \(i\), source-supported information available no later than the
generation cutoff time \(\tau_0\) is organized into two sets:
clinician-visible information \(C_i\) and patient-reportable information
\(P_i\).

The clinician-visible set \(C_i\) contains information available at the
beginning of triage, including selected demographics, recorded triage fields,
initial measurements, and eligible prior-record information. The
patient-reportable set \(P_i\) contains information that may be disclosed by
the patient but is not directly exposed to the clinician, such as symptom
history, associated symptoms, home medications, allergies, and relevant
medical history. A clinical concept may appear in both sets when it is both
documented in the record and independently reportable by the patient.

During generation, the clinician has direct access to \(C_i\), whereas the
patient has direct access to \(P_i\). Both observe the dialogue history, and
the patient additionally receives the non-clinical persona \(\eta_i\).
Post-triage events and the recorded Emergency Severity Index are excluded from
both \(C_i\) and \(P_i\) and cannot condition dialogue generation.

Table~\ref{tab:fact-bank-composition} summarizes the information assigned to
each role. Table~\ref{tab:source-fields} lists the corresponding source fields
and their use in benchmark construction.

\subsection{Role-Based Generation}
\label{sec:dialogue_generation}

Each encounter is additionally associated with a patient persona \(\eta_i\).
The persona specifies non-clinical communication characteristics, such as
verbosity, phrasing style, recall confidence, and the amount of detail
provided in a response. It controls how the patient communicates but does not
add, remove, or modify the clinical facts available to the patient model.

Let
\begin{equation}
H_{i,t-1}
=
\left(
u^c_{i,1},u^p_{i,1},
\ldots,
u^c_{i,t-1},u^p_{i,t-1}
\right)
\label{eq:dialogue-history}
\end{equation}
denote the public dialogue history before round \(t\), where \(u^c_{i,t}\)
and \(u^p_{i,t}\) are the clinician and patient utterances, respectively.

At round \(t\), the clinician generates the next utterance from the
clinician-visible information and the preceding public dialogue:
\begin{equation}
u^c_{i,t}
\sim
p_{\theta_c}
\left(
u
\mid
C_i,
H_{i,t-1}
\right).
\label{eq:clinician-generation}
\end{equation}
The clinician model therefore has no direct access to \(P_i\) and must acquire
patient-reportable information through interaction with the patient.

After the clinician utterance is appended, the intermediate dialogue history
becomes
\begin{equation}
H^c_{i,t}
=
H_{i,t-1}
\oplus
\left(c,u^c_{i,t}\right),
\label{eq:clinician-history-update}
\end{equation}
where \(c\) denotes the clinician role and \(\oplus\) denotes sequence
concatenation.

The patient then generates a response conditioned on the patient-reportable
information, the persona, and the dialogue containing the current clinician
utterance:
\begin{equation}
u^p_{i,t}
\sim
p_{\theta_p}
\left(
u
\mid
P_i,
\eta_i,
H^c_{i,t}
\right).
\label{eq:patient-generation}
\end{equation}
Thus, \(P_i\) constrains the clinical content that the patient may express,
while \(\eta_i\) controls its surface realization.

The completed dialogue round updates the public history as
\begin{equation}
H_{i,t}
=
H_{i,t-1}
\oplus
\left(c,u^c_{i,t}\right)
\oplus
\left(p,u^p_{i,t}\right).
\label{eq:dialogue-history-update}
\end{equation}

Equations~\eqref{eq:clinician-generation}--%
\eqref{eq:dialogue-history-update} define the alternating generation process.
The clinician conditions on \(C_i\) and the public dialogue, whereas the
patient conditions on \(P_i\), the persona \(\eta_i\), and the dialogue
containing the clinician's current question. Accepted disclosures are linked
to their supporting source facts, and the first accepted speaker turn for
each disclosed fact is recorded.

\subsection{Event--Turn Lineage}
\label{sec:verification}

A fluent generated response is not necessarily supported by the source
record. A verifier therefore maps accepted clinical disclosures to normalized
atomic facts and links each fact to its eligible source in \(C_i\) or \(P_i\).

Let \(\Delta Z^{\mathrm{C}}_{i,t}\) and
\(\Delta Z^{\mathrm{P}}_{i,t}\) denote the source-supported facts first
accepted from the clinician and patient, respectively, in round \(t\). The
dialogue-visible information state accumulates as
\begin{equation}
\Delta Z_{i,t}
=
\Delta Z^{\mathrm{C}}_{i,t}
\cup
\Delta Z^{\mathrm{P}}_{i,t},
\qquad
Z_{i,0}
=
\varnothing,
\qquad
Z_{i,t}
=
Z_{i,t-1}
\cup
\Delta Z_{i,t}.
\label{eq:grounded-state}
\end{equation}

Every element of \(\Delta Z_{i,t}\) retains its normalized fact identifier,
supporting EHR source, round index, and speaker role. A completed episode
therefore yields a machine-readable ledger that traces each accepted
disclosure back to its exact source event in the EHR and identifies when and
by whom it first entered the dialogue.

This lineage supports exact measurement of which patient facts were acquired,
when they were acquired, which clinician questions preceded their disclosure,
and whether a disclosed fact originated from an eligible information source.
Only facts originating from \(P_i\) receive information-acquisition credit in
the benchmark metrics. Section~\ref{sec:case-study} provides a complete
annotated example.


\definecolor{ciblock}{HTML}{EAF2FA}
\definecolor{piblock}{HTML}{EDF6EF}

\newcommand{\notapp}{\textit{N/A}}

\begin{table*}[t]
    \centering
    \scriptsize
    \setlength{\tabcolsep}{2.0pt}
    \renewcommand{\arraystretch}{1.13}

    \caption{\textbf{Composition and dialogue realization of source-grounded
    encounter information.}
    Clinician-visible information \(C_i\) is available to the clinician from
    the beginning of the interaction, whereas patient-reportable information
    \(P_i\) must be acquired through dialogue.
    \emph{Encounter coverage} is the percentage of encounters containing at
    least one valid source fact in a category; \emph{mean facts per encounter}
    is the average number of normalized atomic facts.
    The final two columns report the percentage of available facts first
    expressed by each role. Facts that are never expressed are counted toward
    neither column, so the two percentages need not sum to \(100\%\).
    ``N/A'' indicates that a role is not permitted to introduce that category
    as a new disclosure.
    Statistics are computed over all \(4{,}041\) encounters in the frozen
    corpus. Section~\ref{sec:case-study} provides a complete annotated
    example.}
    \label{tab:fact-bank-composition}

    \begin{tabularx}{\textwidth}{
        @{}
        >{\raggedright\arraybackslash}p{0.155\textwidth}
        >{\raggedright\arraybackslash}X
        >{\centering\arraybackslash}p{0.09\textwidth}
        >{\centering\arraybackslash}p{0.09\textwidth}
        >{\centering\arraybackslash}p{0.08\textwidth}
        >{\centering\arraybackslash}p{0.08\textwidth}
        @{}
    }
        \toprule

        Information category
        & Representative example
        & \multicolumn{2}{c}{Source availability}
        & \multicolumn{2}{c}{First introduced by} \\

        \cmidrule(lr){3-4}
        \cmidrule(lr){5-6}

        \multicolumn{2}{
            @{}>{\columncolor{ciblock}}l
        }{
            \shortstack[l]{
                \textbf{Clinician-visible information \(C_i\)}\\[-1pt]
                \textnormal{Available to the clinician at dialogue onset}
            }
        }
        & \shortstack{Encounter\\coverage (\%)}
        & \shortstack{Mean facts\\per encounter}
        & \shortstack{Clinician\\(\%)}
        & \shortstack{Patient\\(\%)} \\

        \midrule

        Demographics
        & Clinician: ``I see that you are 58 years old.''
        & 100.0
        & 1.00
        & 4.8
        & 1.3 \\

        Initial vital signs
        & Clinician: ``I can see your heart rate's a little fast.''
        & 98.0
        & 5.78
        & 95.1
        & 0.0 \\

        Prior conditions
        & Patient: ``I have a history of asthma.''
        & 86.5
        & 3.04
        & 35.7
        & 39.8 \\

        \cmidrule(lr){1-6}
        \addlinespace[3pt]

        \multicolumn{2}{
            @{}>{\columncolor{piblock}}l
        }{
            \shortstack[l]{
                \textbf{Patient-reportable information \(P_i\)}\\[-1pt]
                \textnormal{May be disclosed by the patient through dialogue}
            }
        }
        & & & & \\

        Home medications
        & Patient: ``I take metformin.''
        & 93.5
        & 7.27
        & \notapp
        & 39.7 \\

        Chief complaint
        & Patient: ``I have had nausea since this morning.''
        & 100.0
        & 1.47
        & \notapp
        & 98.8 \\

        Arrival mode
        & Patient: ``I came here by ambulance.''
        & 98.9
        & 0.99
        & \notapp
        & 91.2 \\

        Initial pain
        & Patient: ``The pain is about seven out of ten.''
        & 98.0
        & 0.98
        & \notapp
        & 91.4 \\

        \bottomrule
    \end{tabularx}
\end{table*}

%

%
%

\section{EHR2Dial Validation}
\label{sec:statistics}

We characterize the frozen corpus of \(4{,}041\) encounters and examine
whether the generated dialogues exhibit the properties required by the
benchmark design. Specifically, we assess whether source information is
introduced in accordance with the role-specific access constraints, whether
patient-reportable information is disclosed progressively in the
interaction, and whether the acquired information is predictive of the
recorded triage decision. These analyses evaluate the internal design of the
benchmark; they do not establish clinical validity or readiness for deployment.

\subsection{Role-Consistent Realization of Source Information}
\label{sec:role-consistency}

Table~\ref{tab:fact-bank-composition} summarizes which source facts are
verbalized in the dialogue and which role introduces them first. The observed
realization patterns align with the intended role-specific information access.
Initial vital signs are almost always introduced by the clinician
(\(95.1\%\)), whereas the chief complaint, arrival mode, and initial pain
score are predominantly introduced by the patient
(\(98.8\%\), \(91.2\%\), and \(91.4\%\), respectively). Prior conditions may
be supported by clinician-visible prior records in \(C_i\),
patient-reportable history in \(P_i\), or both, and may therefore first become
explicit through either role.

Not every available source fact needs to be stated explicitly during triage.
Demographic information, for example, is already available to the clinician
and therefore may not be repeated in the conversation. Home medications show a different pattern. Medication records contain an
average of \(7.27\) normalized atomic facts per encounter, of which \(39.7\%\)
are explicitly disclosed. This partial realization may reflect clinician
questioning, the patient response policy, or differences in the relevance of
individual medications to the presenting concern; the realization rate alone
does not distinguish among these explanations. Overall, the dialogues
selectively express available EHR information rather than reproducing the
complete record.

\begin{figure*}[t]
    \centering
    \includegraphics[
        width=\textwidth,
        trim=0 0 0 0,
        clip
    ]{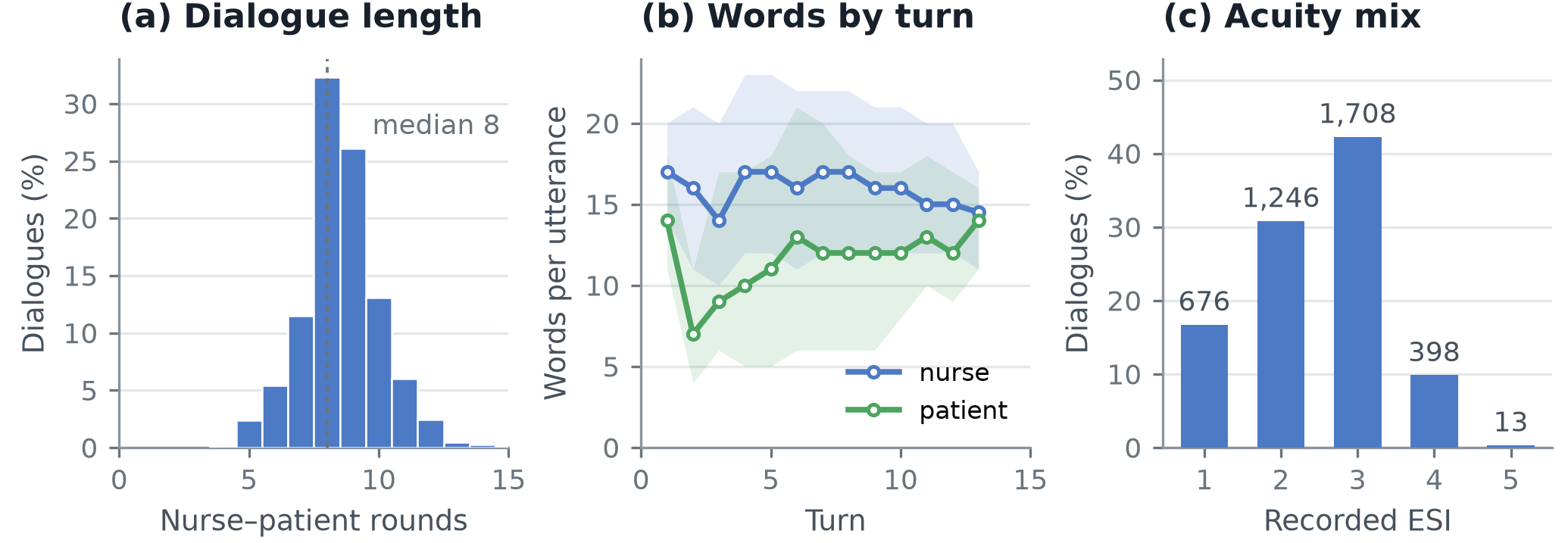}

    \vspace{-0.5em}

    \caption{\textbf{Dialogue structure and acuity composition.}
    Panel (a) shows the distribution of dialogue length in complete
    clinician--patient rounds; the dashed vertical line marks the median of
    eight rounds.
    Panel (b) reports the number of words per clinician and patient utterance
    across rounds, with shaded regions indicating cross-dialogue variation.
    Panel (c) shows the recorded ESI composition of the dialogues included in
    this descriptive analysis; values above the bars indicate the
    corresponding numbers of dialogues.}
    \label{fig:dialogue-shape}
\end{figure*}

\begin{figure*}[t]
    \centering
    \includegraphics[
        width=\textwidth
    ]{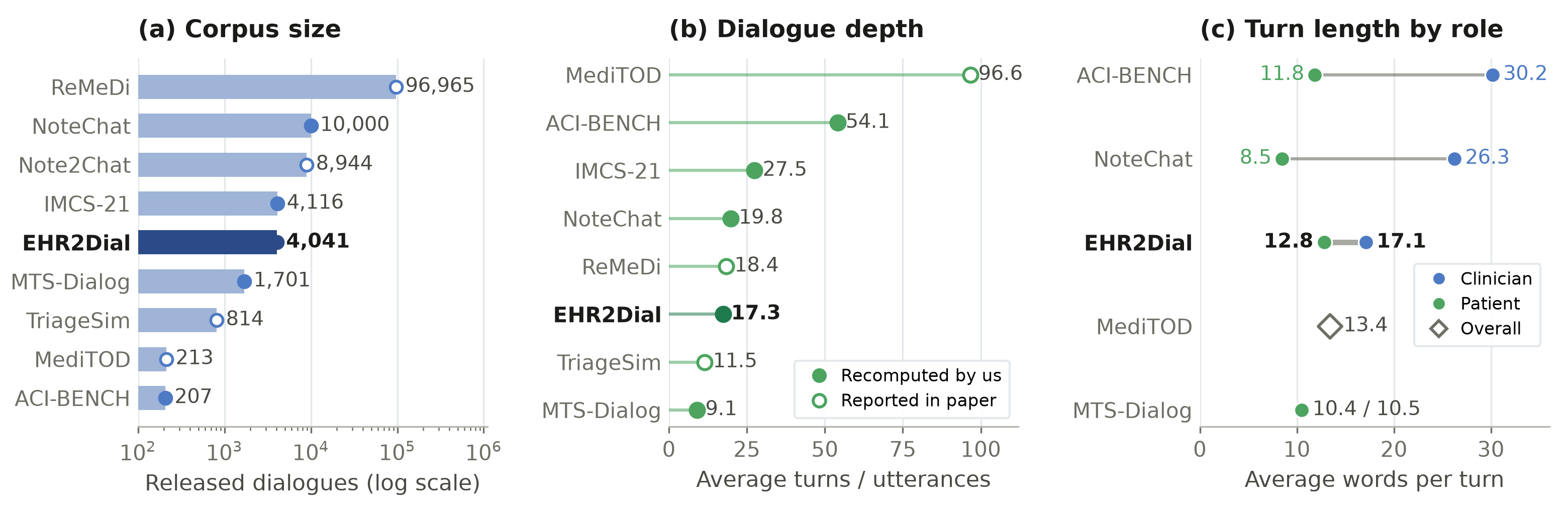}

    \caption{\textbf{Scale and interaction profiles of representative
    clinical dialogue resources.}
    Panel (a) shows the number of released dialogues on a logarithmic scale.
    Panel (b) shows average dialogue depth, and Panel (c) shows the average
    number of words per speaker turn.
    Filled markers denote statistics recomputed from released data, whereas
    hollow markers denote values reported by the original papers;
    unavailable or non-comparable statistics are omitted.
    EHR2Dial-Triage occupies a bounded, variable-length interaction regime:
    its dialogues are sufficiently deep to support multi-turn information
    elicitation while limiting unnecessarily long interactions.
    This constraint is important because an unconstrained clinician model
    could improve terminal information coverage by continuing to ask
    additional questions, even when those questions become repetitive or
    increase interaction burden.}
    \label{fig:corpus-interaction-profile}
\end{figure*}

\subsection{Interaction Scale and Speaker Balance}
\label{sec:interaction-profile}

Figures~\ref{fig:dialogue-shape} and
\ref{fig:corpus-interaction-profile} summarize the scale and interaction
characteristics of EHR2Dial-Triage. The benchmark contains \(4{,}041\)
dialogues, with a median of eight clinician--patient rounds per dialogue.
Most dialogues contain between five and thirteen rounds, providing sufficient
space for multi-turn information elicitation while maintaining a bounded
interaction length.

Clinician utterances are moderately longer than patient utterances.
EHR2Dial-Triage contains an average of \(17.3\) speaker turns per dialogue,
with clinician and patient utterances averaging \(17.1\) and \(12.8\) words,
respectively. As shown in Figure~\ref{fig:dialogue-shape}(b), clinician
utterance length remains relatively stable across rounds, whereas patient
responses generally become longer as the dialogue progresses. In comparison,
several existing clinical dialogue datasets contain substantially longer
clinician utterances than patient responses, producing more asymmetric
interactions.

Interaction length is an important design consideration for evaluating
information-seeking clinical models. A clinician model may collect more
information by asking more or longer questions, which can improve downstream
prediction while increasing repetition and interaction burden. Evaluation
based only on terminal coverage or predictive performance may therefore favor
models that ask more questions rather than models that identify useful
information efficiently. EHR2Dial-Triage uses a bounded, variable-length
interaction budget to limit interview length and support comparable evaluation
across clinician models.

\begin{figure*}[t]
    \centering
    \includegraphics[
        width=0.98\textwidth
    ]{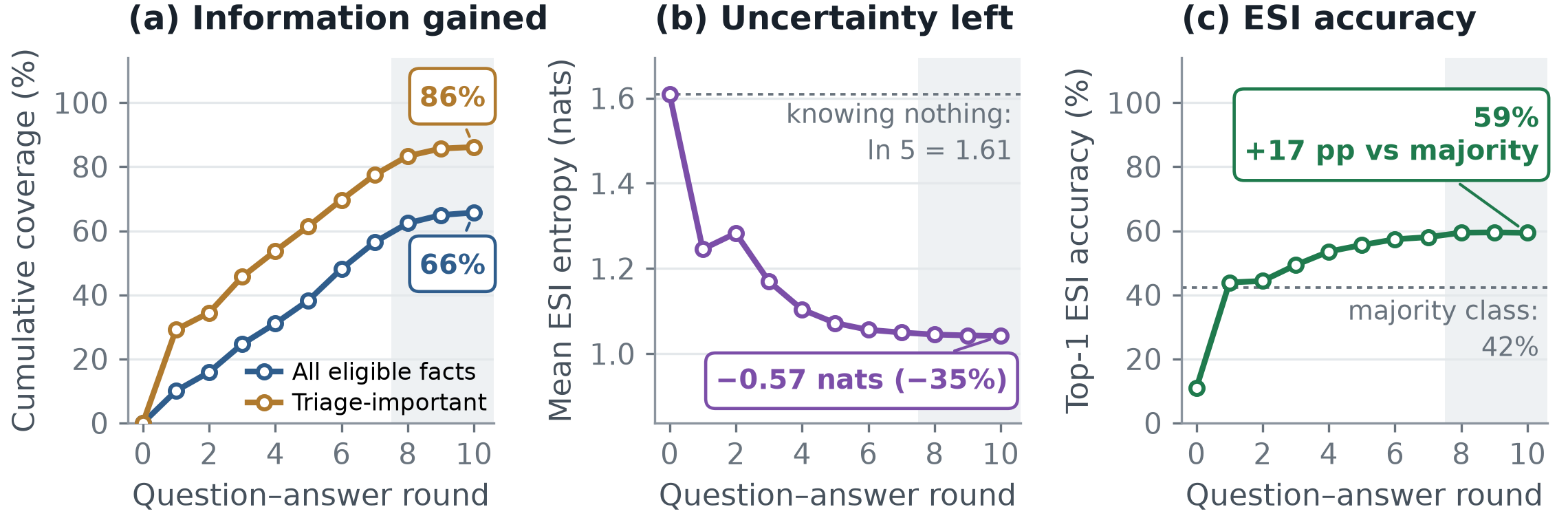}

    \caption{\textbf{Predictive relevance of progressively disclosed
    information.}
    Using event--turn lineage, each accepted disclosure is mapped back to its
    corresponding structured EHR fact. At each question--answer round, a
    fixed, patient-disjoint multinomial logistic regression classifier
    predicts the recorded ESI level from the clinician-visible information
    \(C_i\) and the cumulative patient-reportable facts disclosed up to that
    round.
    Panel (a) shows cumulative coverage of all eligible and triage-important
    patient facts.
    Panel (b) shows mean predictive entropy, where lower values indicate a
    more concentrated predictive distribution.
    Panel (c) shows top-1 recorded-ESI accuracy.
    As additional source-grounded patient information is disclosed,
    predictive entropy decreases and recorded-ESI accuracy increases,
    indicating that the acquired information is predictive of the recorded
    triage label.}
    \label{fig:constructed-dialogue-progression}
\end{figure*}

\subsection{Triage Relevance of Disclosed Information}
\label{sec:information-relevance}

We next examine whether progressively disclosed patient information provides
additional evidence for predicting the recorded ESI level. At each
question--answer round \(t\), we construct a cumulative representation from
\(C_i\) and the patient-reportable facts disclosed by that round. A fixed,
patient-disjoint multinomial logistic regression classifier predicts the
recorded five-level ESI from this representation. Because the classifier is
held fixed across rounds, changes in its predictions reflect changes in the
available patient information.

Figure~\ref{fig:constructed-dialogue-progression} shows that patient information
accumulates progressively over the interaction. By Round~10, the dialogues
cover \(66\%\) of all eligible patient-reportable facts and \(86\%\) of facts
designated as triage-important, indicating that the latter tend to be disclosed
earlier. As additional information becomes available, the reader's predictive
distribution becomes more concentrated and its recorded-ESI accuracy
increases. Mean predictive entropy decreases from \(1.61\) nats initially to
approximately \(1.04\) nats at Round~10, while top-1 ESI accuracy reaches
\(59\%\), compared with a \(42\%\) majority-class baseline.

Overall, these analyses show that EHR2Dial-Triage preserves the intended separation between clinician-visible and patient-reportable information, while allowing patient information to emerge progressively through dialogue. The information disclosed during the interaction is also predictive of the recorded ESI, supporting the use of the benchmark for studying information acquisition in triage. Additional cohort characteristics, including demographic and age distributions, are reported in Appendix~\ref{app:benchmark-cohort-statistics}.
\section{Experiments}
\label{sec:experiments}

\subsection{Experimental Overview}
\label{sec:experimental-overview}

We evaluate LLM capabilities in conversational triage along two complementary
dimensions: information elicitation and downstream triage decision-making.
Experiment~I evaluates LLMs as triage clinicians by examining what questions
they ask, what source-supported patient information they elicit, and how the
acquired evidence changes a fixed ESI reader's assessment. Experiment~II
evaluates whether models can interpret a completed triage dialogue, predict the
recorded five-level Emergency Severity Index (ESI), and generate an appropriate
patient-facing closing. In Experiment~I, each clinician model interacts with
the same patient simulator, whereas Experiment~II provides all models with the
same completed dialogue. This design separates the ability to acquire
clinically relevant information from the ability to use and communicate that
information. Appendix~\ref{app:evaluation-details} provides the complete metric
definitions and aggregation procedures.


\subsection{Experiment I: Clinician Information Elicitation}
\label{sec:exp-active-elicitation}

We examine what questions different LLMs ask when acting as triage clinicians. Each model interacts with the same patient simulator, allowing us to compare turn-level questioning patterns and the information elicited across models.

\paragraph{Setup.}
For encounter \(i\), the clinician model observes the clinician-visible
information \(C_i\) and the preceding dialogue. Eligible patient-reportable
facts, the recorded ESI label, and post-triage events are not shown to the
clinician model. The patient agent, fact bank, disclosure policy, verifier,
stopping rule, and ten-round budget are fixed across models.

One round consists of a clinician question followed by a patient response. A
fact is counted when it first appears in an accepted patient response.
Repeated disclosures receive no additional credit, and facts already
available in \(C_i\) are excluded from the acquisition metrics.

\paragraph{Metrics.}
We assign each clinician question one primary information target and report
the distribution of targets for each model. At rounds
\(t\in\{2,6,10\}\), \textsc{FactCov}@\(t\) measures the cumulative
proportion of eligible patient-reportable facts that have been disclosed.
\textsc{HistoryMedCov}@\(t\) applies the same calculation to medical-history
and home-medication facts. Both metrics are computed within each encounter
and then macro-averaged. We also apply a fixed, patient-disjoint multinomial logistic-regression reader
to \(C_i\) and the cumulative acquired facts at each checkpoint. We report
the mean entropy of its five-class ESI distribution in nats; lower values
indicate a more concentrated reader distribution.
Appendix~\ref{app:evaluation-details} provides the full definitions.

\begin{table*}[t]
    \centering
    \scriptsize
    \setlength{\tabcolsep}{3.0pt}
    \renewcommand{\arraystretch}{1.10}

    \caption{\textbf{Turn-wise fact acquisition and reader entropy.}
    \textsc{FactCov} reports cumulative coverage of all eligible
    patient-reportable facts, and \textsc{HistoryMedCov} reports coverage of
    eligible medical-history and home-medication facts. Entropy is computed
    from the fixed five-level ESI reader at each checkpoint. Coverage values
    are percentages. All models are evaluated on the same
    encounter set. Best open-weight results are bold, with ties retained.}
    \label{tab:exp1-complete-main}

    \begin{adjustbox}{max width=\textwidth}
    \begin{tabular}{@{}lccccccccc@{}}
        \toprule
        & \multicolumn{3}{c}{\textsc{FactCov} (All Facts)}
        & \multicolumn{3}{c}{\textsc{HistoryMedCov} (His\&Med)}
        & \multicolumn{3}{c}{Reader Uncertainty} \\
        \cmidrule(lr){2-4}
        \cmidrule(lr){5-7}
        \cmidrule(lr){8-10}

        Clinician model
        & \shortstack{Cov@2\\\(\uparrow\)}
        & \shortstack{Cov@6\\\(\uparrow\)}
        & \shortstack{Cov@10\\\(\uparrow\)}
        & \shortstack{Cov@2\\\(\uparrow\)}
        & \shortstack{Cov@6\\\(\uparrow\)}
        & \shortstack{Cov@10\\\(\uparrow\)}
        & \shortstack{Entropy@2\\\(\downarrow\)}
        & \shortstack{Entropy@6\\\(\downarrow\)}
        & \shortstack{Entropy@10\\\(\downarrow\)} \\
        \midrule

        \multicolumn{10}{@{}l}{\textit{General-purpose open-weight models}} \\

        Qwen3-14B-AWQ~\citep{yang2025qwen3,lin2024awq}
        & 19.2 & 52.8 & 60.0
        & 8.4 & 37.8 & 45.2
        & 1.1097 & 0.8056 & 0.7510 \\

        GLM-4.7-Flash~\citep{zai2026glm47flash,glmteam2025glm45}
        & 19.7 & \textbf{53.9} & \textbf{68.0}
        & 8.9 & 38.7 & \textbf{56.5}
        & 1.1036 & \textbf{0.7915} & 0.6927 \\

        gpt-oss-20B~\citep{openai2025gptoss}
        & 19.5 & 53.4 & 66.9
        & 9.0 & 38.5 & 55.0
        & 1.1030 & 0.7949 & \textbf{0.6923} \\

        Gemma-4-12B~\citep{gemmateam2026gemma4}
        & 18.9 & 52.7 & 58.1
        & 7.4 & 37.8 & 43.1
        & 1.1158 & 0.8108 & 0.7625 \\

        Qwen3.5-4B~\citep{qwen2026qwen35}
        & 19.2 & 53.2 & 60.3
        & 8.4 & 38.5 & 45.8
        & 1.1143 & 0.8079 & 0.7487 \\

        Qwen3.5-9B~\citep{qwen2026qwen35}
        & 19.2 & 53.1 & 64.1
        & 7.9 & 38.0 & 50.9
        & 1.1027 & 0.7988 & 0.7133 \\

        Qwen3.6-27B~\citep{qwen2026qwen3627b}
        & 19.0 & 52.9 & 59.6
        & 7.9 & 37.9 & 44.6
        & 1.1167 & 0.7989 & 0.7485 \\

        Qwen3.6-35B-A3B~\citep{qwen2026qwen3635ba3b}
        & 19.0 & 52.8 & 58.1
        & 8.0 & 37.9 & 43.2
        & 1.1098 & 0.8001 & 0.7541 \\

        \midrule
        \multicolumn{10}{@{}l}{\textit{Domain-specialized open-weight model}} \\

        MedGemma-1.5-4B~\citep{sellergren2026medgemma15}
        & \textbf{19.8} & \textbf{53.9} & 67.7
        & \textbf{9.7} & \textbf{38.9} & 55.9
        & \textbf{1.0999} & \textbf{0.7915} & 0.6957 \\

        \midrule
        \multicolumn{10}{@{}l}{\textit{Proprietary frontier models}} \\

        GPT-5.6~\citep{openai2026gpt56}
        & 19.3 & 53.7 & 66.1
        & 8.1 & 38.7 & 53.7
        & 1.0184 & 0.8112 & 0.7497 \\

        Claude Fable 5~\citep{anthropic2026fable5}
        & 19.4 & 53.9 & 63.9
        & 8.3 & 38.9 & 50.6
        & 1.0215 & 0.8096 & 0.7528 \\

        \bottomrule
    \end{tabular}
    \end{adjustbox}
\end{table*}

\begin{figure*}[t]
    \centering
    \includegraphics[width=\textwidth]{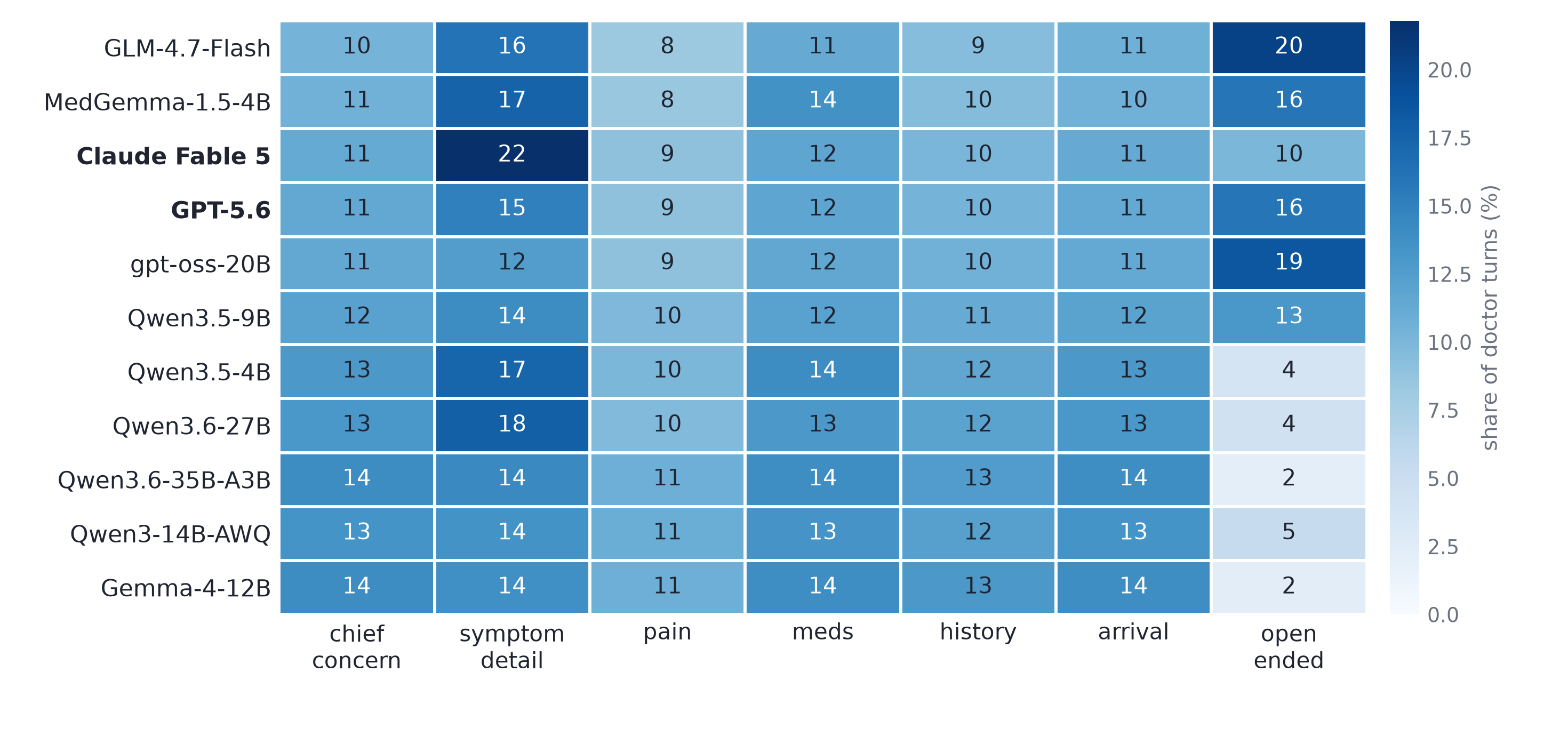}
\caption{\textbf{Distribution of clinician questions across primary information targets.}
For each clinician model, every valid question is assigned to the single clinical
information category that best represents its primary intent, such as presenting
complaint, symptom characteristics, pain, medical history, medications, or
open-ended follow-up. Each cell shows the percentage of that model's questions
assigned to the corresponding category, so each row sums to \(100\%\).
The figure compares how different models distribute their questioning
effort across clinical topics during the triage interaction.}
    \label{fig:question-targets}
\end{figure*}

\paragraph{Information acquisition.}
Table~\ref{tab:exp1-complete-main} reports turn-wise information coverage and
ESI-reader uncertainty across clinician models. Models behave similarly during the early part of the interaction. At Round~2, FactCov ranges only from \(18.9\%\) to \(19.8\%\), and coverage remains tightly clustered at Round~6 \(52.7\%\)--\(53.9\%\). This suggests that most models identify a similar set of high-priority questions early in the interview, when common symptom, history, and medication information is still readily available. MedGemma-1.5-4B performs well in this early stage, achieving the highest FactCov@2 and HistoryMedCov@2 and remaining at or tied for the highest coverage at Round~6. One possible explanation is that medical pretraining helps the model prioritize clinically salient follow-up questions early in the conversation. The advantage is modest, however, and the early trajectories are otherwise similar across model families.

\paragraph{Question allocation.}
Figure~\ref{fig:question-targets} provides a clearer view of how the models differ in their questioning behavior. Across models, most questions concern the presenting complaint, symptom characteristics, and pain, but the relative emphasis on these domains varies. The largest difference appears in symptom-detail questions, which account for roughly \(12\%\)--\(22\%\) of all questions. MedGemma-1.5-4B and several strong general-purpose models devote a larger share of their questions to clarifying symptom characteristics, suggesting a stronger emphasis on refining the clinical picture before moving to other topics.

Models also differ in how narrowly they formulate follow-up questions. GPT-5.6, GLM-4.7-Flash, and MedGemma-1.5-4B make greater use of open-ended follow-ups, whereas other models rely more heavily on targeted questions tied to a specific clinical domain. These strategies represent different ways of eliciting information: targeted questions are more controlled, while open-ended questions may allow patients to introduce information that the clinician did not explicitly anticipate.

\paragraph{Later-round acquisition and reader uncertainty.}
The models begin to diverge more clearly in the later rounds. By Round~10, FactCov ranges from \(58.1\%\) to \(68.0\%\), and HistoryMedCov from \(43.1\%\) to \(56.5\%\). GLM-4.7-Flash reaches the highest terminal coverage, followed closely by MedGemma-1.5-4B and gpt-oss-20B. This pattern suggests that the main differences across models emerge after the most obvious information has already been collected, when further progress depends on whether the model continues to identify additional relevant topics or invites the patient to provide information beyond the immediately targeted question.

Models that combine targeted symptom questions with open-ended follow-ups tend to achieve higher terminal coverage, although the question-target distribution alone does not establish a causal relationship. Reader entropy also decreases for every clinician model as information accumulates, indicating that the elicited facts generally make the downstream ESI assessment more concentrated. At Round~10, gpt-oss-20B, GLM-4.7-Flash, and MedGemma-1.5-4B produce the lowest reader entropy. Their entropy values are similar despite differences in total coverage, suggesting that the value of the acquired information depends not only on how many facts are elicited, but also on which facts are obtained.


\subsection{Experiment II: Triage Prediction and Closing Generation}
\label{sec:exp-dialogue-understanding}

We next evaluate whether LLMs can use the information contained in a completed triage dialogue to predict the patient's ESI level and provide a clear patient-facing explanation of the assigned acuity and its clinical rationale.

\paragraph{Setup.}
For each encounter \(i\), every model receives the same completed,
speaker-labeled dialogue \(\widetilde{H}_i\). The prompt does not expose the
recorded ESI label, benchmark reference closing, separately serialized EHR
fields, source identifiers, provenance annotations, verifier decisions, or
post-triage information. Under a common output schema, each model returns a
five-class score vector, a discrete ESI prediction, and a concise
patient-facing closing. The score vector is used to compute Macro-AUC and the
score-based error analysis below.

\paragraph{Metrics.}
For ESI prediction, we report Macro-AUC, Macro-F1, quadratic weighted kappa
(QWK), and under-triage rate. Macro-AUC averages one-vs-rest discrimination
over the five ESI levels, Macro-F1 weights all levels equally, and QWK accounts
for the ordinal distance between predicted and recorded ESI
levels~\citep{cohen1968weighted}. Since larger ESI values indicate lower
urgency, under-triage is defined as \(\hat{y}_i>y_i\).

For closing generation, \textsc{Key Recall} measures coverage of important
dialogue-supported units, and \textsc{Grounded Precision} measures the
proportion of generated clinical claims supported by the dialogue. We also
report BERTScore F1~\citep{zhang2020bertscore},
ROUGE-L~\citep{lin-2004-rouge}, BLEU-4~\citep{papineni-etal-2002-bleu}, and
corpus-level Distinct-2~\citep{li-etal-2016-diversity}.
Appendix~\ref{app:evaluation-details} provides the full definitions.

\begin{table*}[t]
\centering
\scriptsize
\setlength{\tabcolsep}{2.5pt}
\renewcommand{\arraystretch}{1.10}

```
\caption{\textbf{Dialogue-based ESI prediction and patient-facing closing
generation.}
Every model receives the same completed dialogue. Macro-AUC, Macro-F1, QWK, and
under-triage evaluate ESI prediction. \textsc{Key Recall} and
\textsc{Grounded Precision} evaluate closing content and support;
BERTScore F1, ROUGE-L, and BLEU-4 measure reference similarity; and
Distinct-2 measures corpus-level lexical diversity. Best displayed results
are bold.}
\label{tab:dialogue-esi-and-closing}

\begin{adjustbox}{max width=\textwidth}
\begin{tabular}{@{}lcccccccccc@{}}
    \toprule
    & \multicolumn{4}{c}{ESI prediction}
    & \multicolumn{6}{c}{Patient-facing closing generation} \\
    \cmidrule(lr){2-5}
    \cmidrule(lr){6-11}

    & \multicolumn{4}{c}{}
    & \multicolumn{2}{c}{Source grounding}
    & \multicolumn{3}{c}{Reference similarity}
    & \multicolumn{1}{c}{Diversity} \\
    \cmidrule(lr){6-7}
    \cmidrule(lr){8-10}
    \cmidrule(lr){11-11}

    Model
    & \shortstack{Macro-\\AUC \(\uparrow\)}
    & \shortstack{Macro-\\F1 \(\uparrow\)}
    & QWK \(\uparrow\)
    & Under \(\downarrow\)
    & \shortstack{Key\\Recall \(\uparrow\)}
    & \shortstack{Grounded\\Prec. \(\uparrow\)}
    & \shortstack{BERTScore\\F1 \(\uparrow\)}
    & \shortstack{ROUGE-L\\\(\uparrow\)}
    & \shortstack{BLEU-4\\\(\uparrow\)}
    & \shortstack{Dist-2\\\(\uparrow\)} \\
    \midrule

    \multicolumn{11}{@{}l}{\textit{General-purpose open-weight models}} \\

    Qwen3-14B-AWQ~\citep{yang2025qwen3,lin2024awq}
    & 0.574 & 0.173 & 0.123 & 0.471
    & 0.376 & 0.406 & 0.234 & 0.155 & 0.051 & 0.149 \\

    GLM-4.7-Flash~\citep{zai2026glm47flash,glmteam2025glm45}
    & 0.559 & 0.179 & 0.052 & 0.621
    & 0.284 & 0.445 & 0.252 & 0.157 & \textbf{0.053} & 0.148 \\

    gpt-oss-20B~\citep{openai2025gptoss}
    & 0.611 & 0.221 & 0.321 & 0.531
    & 0.398 & 0.425 & 0.202 & 0.150 & 0.048 & 0.165 \\

    Gemma-4-12B~\citep{gemmateam2026gemma4}
    & 0.668 & 0.253 & 0.404 & 0.453
    & 0.366 & 0.470 & 0.247 & \textbf{0.160} & 0.052 & 0.131 \\

    Qwen3.5-4B~\citep{qwen2026qwen35}
    & 0.517 & 0.198 & 0.159 & 0.530
    & 0.394 & 0.466 & 0.229 & 0.153 & 0.045 & 0.144 \\

    Qwen3.5-9B~\citep{qwen2026qwen35}
    & 0.630 & 0.243 & 0.298 & \textbf{0.337}
    & 0.450 & 0.461 & 0.212 & 0.149 & 0.044 & 0.140 \\

    Qwen3.6-27B~\citep{qwen2026qwen3627b}
    & 0.668 & 0.297 & 0.367 & 0.440
    & 0.437 & 0.451 & 0.212 & 0.154 & 0.046 & 0.134 \\

    Qwen3.6-35B-A3B~\citep{qwen2026qwen3635ba3b}
    & 0.671 & 0.293 & 0.411 & 0.460
    & 0.406 & 0.464 & 0.208 & 0.150 & 0.044 & 0.133 \\

    \midrule
    \multicolumn{11}{@{}l}{\textit{Domain-specialized open-weight model}} \\

    MedGemma-1.5-4B~\citep{sellergren2026medgemma15}
    & 0.621 & 0.135 & 0.124 & 0.405
    & 0.269 & 0.311 & \textbf{0.263} & 0.146 & 0.051 & 0.122 \\

    \midrule
    \multicolumn{11}{@{}l}{\textit{Proprietary frontier models}} \\

    GPT-5.6 (\texttt{gpt-5.6-sol})~\citep{openai2026gpt56}
    & 0.711 & 0.277 & 0.400 & 0.373
    & \textbf{0.562} & \textbf{0.660} & 0.178
    & 0.149 & 0.049 & \textbf{0.185} \\

    Claude Fable 5~\citep{anthropic2026fable5}
    & \textbf{0.763} & \textbf{0.321} & \textbf{0.517} & 0.413
    & 0.385 & 0.443 & 0.217 & 0.156 & 0.042 & 0.131 \\

    \bottomrule
\end{tabular}
\end{adjustbox}
```

\end{table*}

\paragraph{Results.}
Table~\ref{tab:dialogue-esi-and-closing} reports two complementary evaluations:
ESI prediction and patient-facing language generation. Because all models
receive the same completed dialogue, the comparison reflects how they use the
same available information.

\paragraph{Triage prediction.}
Claude Fable~5 performs best overall on ESI prediction, with the highest
Macro-AUC, Macro-F1, and QWK. This indicates stronger performance in both
distinguishing acuity levels and matching the recorded ordinal ESI labels.
Among the open-weight models, Qwen3.6-35B-A3B achieves the highest Macro-AUC
and QWK, while Qwen3.6-27B obtains the highest Macro-F1. The different rankings
across metrics suggest that separating acuity levels and assigning the exact
recorded ESI are not identical tasks.

\paragraph{Patient-facing generation.}
GPT-5.6 performs best on the grounding-oriented generation metrics, with the
highest Key Recall and Grounded Precision. Its responses therefore include more
of the designated key information while remaining better supported by the
dialogue. GPT-5.6 also achieves the highest Distinct-2, indicating greater
lexical diversity across generated closings. Reference-similarity metrics produce a different ranking. MedGemma-1.5-4B
achieves the highest BERTScore F1, Gemma-4-12B the highest ROUGE-L, and
GLM-4.7-Flash the highest BLEU-4. These metrics measure similarity to the
reference response, whereas Key Recall and Grounded Precision focus more
directly on information coverage and dialogue support.


\begin{figure*}[t]
    \centering
    \includegraphics[width=\textwidth]{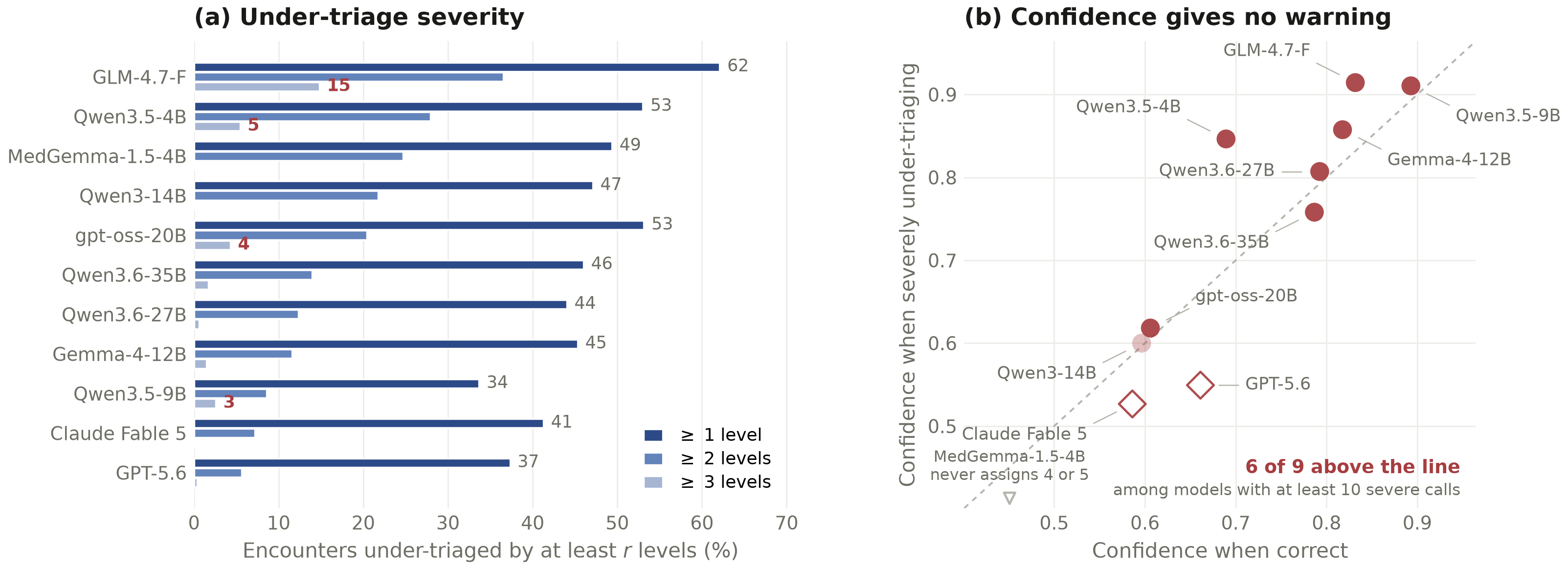}
    \caption{\textbf{Under-triage severity and the absence of a confidence
    signal.} \textbf{(a)} Cumulative under-triage rate
    $U^{(r)}=\Pr(\hat y-y\ge r)$ ;
    larger predicted ESI values denote lower assigned urgency, so $r$ counts
    levels of missed acuity. The three bars per model are nested by
    construction. Numbers are printed for $r=1$ and, where it exceeds $2\%$,
    for $r=3$. \textbf{(b)} Mean maximum reported class probability on
    correct predictions against the same model's mean on \emph{severe}
    under-triage, defined as assigning ESI~4 or~5 to an encounter recorded
    as ESI~1 or~2. The dashed diagonal marks equal confidence; a model above
    it reports more confidence when dangerously wrong than when right.
    Diamonds denote proprietary models. Faded markers rest on fewer than ten
    severe calls and are excluded from the count. MedGemma-1.5-4B never
    assigns ESI~4 or~5, so no severe-under-triage confidence exists; it is
    drawn on the floor at its correct-call mean of $0.451$.}
    \label{fig:safety-undertriage}
\end{figure*}

\subsection{Error Analysis: Under-Triage and Reported Confidence}
\label{sec:safety-analysis}

Under-triage is clinically important because assigning a patient to a lower
urgency level may delay evaluation or treatment. We therefore examine not
only how often under-triage occurs, but also how large these errors are.
Figure~\ref{fig:safety-undertriage}(a) reports the cumulative under-triage
rate
\[
U^{(r)}=\Pr(\hat y-y\ge r),
\]
for \(r\in\{1,2,3\}\). Model rankings change across these thresholds,
showing that frequent under-triage does not always correspond to larger
errors. GPT-5.6 and Claude Fable~5 show lower rates of multi-level
under-triage than several open-weight models, highlighting the importance of
examining error severity in addition to overall error frequency.

Figure~\ref{fig:safety-undertriage}(b) tests whether reported confidence
provides a warning when severe under-triage occurs. Severe under-triage is
defined as predicting ESI~4 or~5 for an encounter recorded as ESI~1 or~2.
For most models with enough severe cases for comparison, confidence on these
errors is similar to or higher than confidence on correct predictions.
GPT-5.6 and Claude Fable~5 show the opposite pattern, with lower confidence
when severe under-triage occurs.

These results raise an important safety concern. A model can make a large
triage error without showing a corresponding drop in confidence, making such
failures difficult to detect from reported scores alone. Safety evaluation
for interactive triage systems should therefore consider the severity of
under-triage, the reliability of confidence estimates, and mechanisms for
escalation or human review when uncertainty or risk is high.

\section{Conclusion}
\label{sec:conclusion}

We introduced \textbf{EHR2Dial-Triage}, a source-grounded benchmark for
studying multi-turn information acquisition and decision making in
emergency-department triage. Built from MIMIC-IV-ED encounters, the framework
controls information availability during the interaction and traces
source-supported facts as they enter the dialogue. Across 4,041 conversations,
our experiments show that information acquisition, ESI prediction, evidence
grounding, and patient-facing communication capture distinct dimensions of
model performance. Models that perform strongly on final acuity prediction do
not necessarily achieve the strongest information acquisition or grounded
communication, highlighting the value of evaluating conversational triage as a
multi-stage clinical process.

These results also point to broader considerations for the safety of
interactive medical AI. Safe clinical dialogue depends on more than the final
prediction: a system must identify missing information, elicit clinically
relevant evidence, recognize changes in urgency, and communicate the immediate
plan clearly to the patient. EHR2Dial-Triage provides a framework for studying
these behaviors with explicit source and temporal grounding. Future work can
extend this setting through clinician-adjudicated safety evaluation,
multi-institutional validation, and richer assessment of patient-facing
communication, including clarity, actionability, and responses to evolving or
ambiguous patient reports. More broadly, evaluating how medical AI acquires,
uses, and communicates clinical information will be important as conversational
systems become increasingly integrated into patient-facing care.


\bibliographystyle{iclr2025_conference}
\bibliography{references}



\appendix


\definecolor{CasePurple}{HTML}{6F6385}
\definecolor{CaseLavender}{HTML}{F4F0F8}
\definecolor{CaseLavenderBorder}{HTML}{D8CEE3}
\definecolor{CaseText}{HTML}{2F3A40}

\definecolor{ClinicianBlue}{HTML}{7FAFCC}
\definecolor{ClinicianFill}{HTML}{ECF5FA}
\definecolor{ClinicianBorder}{HTML}{BFD9E8}
\definecolor{ClinicianText}{HTML}{3E7293}

\definecolor{PatientGreen}{HTML}{8DB99A}
\definecolor{PatientFill}{HTML}{EFF7F1}
\definecolor{PatientBorder}{HTML}{C5DDCB}
\definecolor{PatientText}{HTML}{4F7F5E}

\definecolor{GroundGold}{HTML}{94784D}


\newcommand{\clinicianavatar}{%
    \includegraphics[
        width=7mm,
        height=7.5mm,
        keepaspectratio
    ]{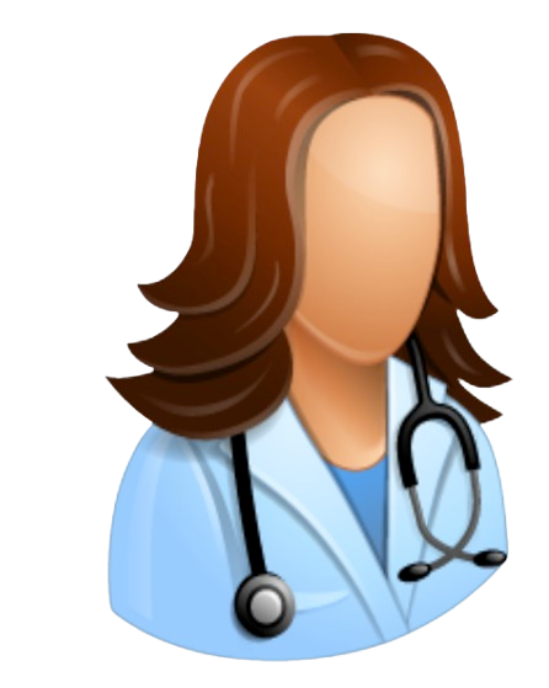}%
}

\newcommand{\patientavatar}{%
    \includegraphics[
        width=7mm,
        height=7.5mm,
        keepaspectratio
    ]{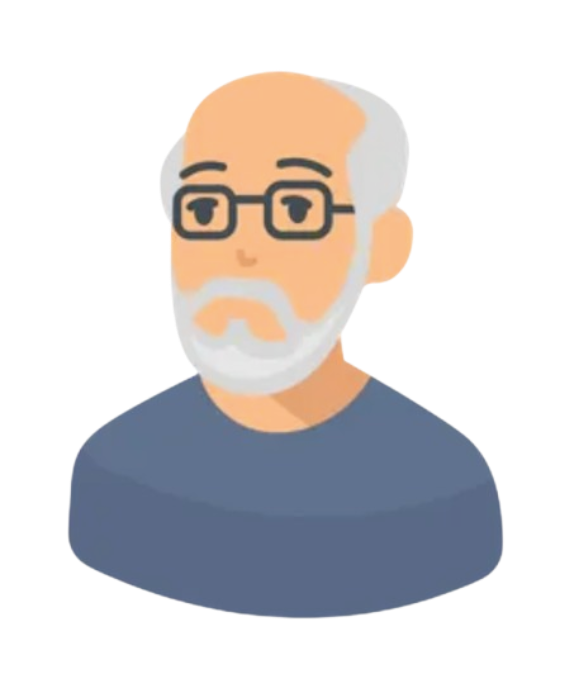}%
}


\newtcolorbox{dialoguecase}{
    enhanced,
    breakable,
    colback=white,
    colframe=CaseLavenderBorder,
    colbacktitle=CasePurple,
    coltitle=white,
    coltext=CaseText,
    fonttitle=\sffamily\bfseries\small,
    title={A Grounded Triage Conversation},
    toptitle=4pt,
    bottomtitle=4pt,
    boxrule=0.55pt,
    arc=1.5mm,
    outer arc=1.5mm,
    borderline west={2.2pt}{0pt}{CasePurple},
    boxsep=0pt,
    left=7pt,
    right=7pt,
    top=7pt,
    bottom=7pt,
    before skip=6pt,
    after skip=6pt,
    width=\columnwidth,
    fontupper=\small
}


\newtcolorbox{arrivalcontext}{
    enhanced,
    colback=CaseLavender,
    colframe=CaseLavenderBorder,
    boxrule=0.4pt,
    arc=1mm,
    borderline west={2pt}{0pt}{CasePurple!75},
    left=6pt,
    right=6pt,
    top=6pt,
    bottom=6pt,
    before skip=1pt,
    after skip=7pt
}


\newtcolorbox{clinicianbubblebox}{
    enhanced,
    width=\linewidth,
    colback=ClinicianFill,
    colframe=ClinicianBorder,
    boxrule=0.4pt,
    arc=2mm,
    outer arc=2mm,
    borderline west={2pt}{0pt}{ClinicianBlue},
    left=7pt,
    right=7pt,
    top=5pt,
    bottom=5pt,
    before skip=0pt,
    after skip=0pt
}

\newtcolorbox{patientbubblebox}{
    enhanced,
    width=\linewidth,
    colback=PatientFill,
    colframe=PatientBorder,
    boxrule=0.4pt,
    arc=2mm,
    outer arc=2mm,
    borderline east={2pt}{0pt}{PatientGreen},
    left=7pt,
    right=7pt,
    top=5pt,
    bottom=5pt,
    before skip=0pt,
    after skip=0pt
}


\newtcolorbox{closingstrip}{
    enhanced,
    colback=CaseLavender,
    colframe=CaseLavenderBorder,
    boxrule=0.45pt,
    arc=1.2mm,
    borderline west={2pt}{0pt}{CasePurple},
    left=7pt,
    right=7pt,
    top=6pt,
    bottom=6pt,
    before skip=7pt,
    after skip=2pt
}


\newcommand{\turnbadge}[1]{%
    \par\medskip
    \noindent
    \makebox[\linewidth][c]{%
        \tcbox[
            on line,
            colback=CasePurple,
            colframe=CasePurple,
            boxrule=0pt,
            arc=0.8mm,
            left=4pt,
            right=4pt,
            top=1pt,
            bottom=1pt
        ]{%
            \color{white}
            \sffamily\bfseries\scriptsize
            TURN~#1
        }%
    }
    \par\smallskip
}


\newlength{\dialoguebubblewidth}

\newcommand{\fitbubblewidth}[1]{%
    \settowidth{\dialoguebubblewidth}{\small #1}%
    \addtolength{\dialoguebubblewidth}{26pt}%
    \ifdim\dialoguebubblewidth<0.30\linewidth
        \setlength{\dialoguebubblewidth}{0.30\linewidth}%
    \fi
    \ifdim\dialoguebubblewidth>0.78\linewidth
        \setlength{\dialoguebubblewidth}{0.78\linewidth}%
    \fi
}


\newcommand{\clinicianbubble}[1]{%
    \par\noindent
    \begingroup
    \fitbubblewidth{#1}%
    \begin{minipage}[t]{0.075\linewidth}
        \centering
        \vspace{3pt}
        \clinicianavatar
    \end{minipage}%
    \hspace{0.012\linewidth}%
    \begin{minipage}[t]{\dialoguebubblewidth}
        \begin{clinicianbubblebox}
            {\scriptsize\sffamily\bfseries
            \color{ClinicianText}CLINICIAN}\par
            \smallskip
            #1
        \end{clinicianbubblebox}
    \end{minipage}
    \endgroup
    \par\smallskip
}


\newcommand{\patientbubble}[1]{%
    \par\noindent
    \begingroup
    \fitbubblewidth{#1}%
    \hfill
    \begin{minipage}[t]{\dialoguebubblewidth}
        \begin{patientbubblebox}
            \hfill
            {\scriptsize\sffamily\bfseries
            \color{PatientText}PATIENT}\par
            \smallskip
            \raggedleft
            #1
        \end{patientbubblebox}
    \end{minipage}%
    \hspace{0.012\linewidth}%
    \begin{minipage}[t]{0.075\linewidth}
        \centering
        \vspace{3pt}
        \patientavatar
    \end{minipage}
    \endgroup
    \par\smallskip
}


\newcommand{\groundingnote}[1]{%
    \noindent
    \hfill
    \begin{minipage}{0.76\linewidth}
        {\scriptsize\sffamily
        \textcolor{GroundGold}{\faLink\ \bfseries Grounding.}
        #1}
    \end{minipage}
    \hspace*{0.087\linewidth}
    \par\smallskip
}


\newcommand{\caseheading}[1]{%
    \par\medskip
    \noindent
    {\sffamily\bfseries\color{CasePurple}#1}
    \par\smallskip
}


\section{Case Study}
\label{sec:case-study}

This example shows how source-grounded information enters a triage
conversation through either the clinician or the patient. One turn contains one
clinician utterance and one patient response. The annotations identify who first
makes each fact explicit in the conversation and link that fact to its
supporting record.

\begin{dialoguecase}


\begin{arrivalcontext}

The patient presents with \emph{nausea and abdominal discomfort}.
The recorded pain score is \(6/10\), with heart rate
\(102\) beats/min, respiratory rate \(18\) breaths/min,
blood pressure \(148/86\) mmHg, and oxygen saturation \(98\%\).

\smallskip

{\scriptsize\sffamily\color{CasePurple!85}
Source:
\texttt{triage.chiefcomplaint},
\texttt{triage.pain},
\texttt{triage.heartrate},
\texttt{triage.resprate},
\texttt{triage.sbp},
\texttt{triage.dbp}, and
\texttt{triage.o2sat}.}

\end{arrivalcontext}


\turnbadge{1}

\clinicianbubble{
Hi, I am your triage clinician. I am sorry you are feeling so uncomfortable.
Can you tell me where the discomfort is and what it feels like?
}

\patientbubble{
It is in the middle of my stomach. It feels like a constant ache, and I am
also feeling nauseated.
}

\groundingnote{
The patient reveals the nausea and abdominal discomfort recorded in
\texttt{triage.chiefcomplaint}.
}


\turnbadge{2}

\clinicianbubble{
Thank you. Your heart rate is a little fast at 102. On a scale from zero to
ten, how bad is the pain right now?
}

\patientbubble{
It is about a six out of ten.
}

\groundingnote{
The clinician makes \texttt{triage.heartrate = 102} explicit, and the patient
reports the recorded pain score \texttt{triage.pain = 6} in response to the
clinician's question.
}


\turnbadge{3}

\clinicianbubble{
Do you have any ongoing health conditions?
}

\patientbubble{
I have high blood pressure and type 2 diabetes.
}

\groundingnote{
The patient reveals two source-supported history facts aligned with
prior-record diagnosis entries for hypertension (\texttt{I10}) and type 2
diabetes (\texttt{E119}, normalized as E11.9).
}


\turnbadge{4}

\clinicianbubble{
What medicines do you regularly take at home?
}

\patientbubble{
I take metformin for diabetes, lisinopril for my blood pressure, and aspirin.
}

\groundingnote{
The patient reveals the medication facts supported by
\texttt{medrecon.name = METFORMIN},
\texttt{medrecon.name = LISINOPRIL}, and
\texttt{medrecon.name = ASPIRIN}. No unsupported dose or frequency is added.
}


\caseheading{\faListAlt\enspace Information revealed in the conversation}

{\scriptsize
\renewcommand{\arraystretch}{1.18}

\begin{tabularx}{\linewidth}{
    @{}
    >{\centering\arraybackslash}p{0.07\linewidth}
    >{\raggedright\arraybackslash}p{0.17\linewidth}
    >{\raggedright\arraybackslash}p{0.34\linewidth}
    >{\raggedright\arraybackslash}X
    @{}
}
\toprule

Turn
& Revealed by
& Information made explicit
& Supporting record \\

\midrule

1
& \textcolor{PatientText}{\sffamily\bfseries Patient}
& Nausea; abdominal discomfort
& \texttt{triage.chiefcomplaint} \\

2
& \textcolor{ClinicianText}{\sffamily\bfseries Clinician}
& Heart rate 102
& \texttt{triage.heartrate} \\

2
& \textcolor{PatientText}{\sffamily\bfseries Patient}
& Current pain 6/10
& \texttt{triage.pain} \\

3
& \textcolor{PatientText}{\sffamily\bfseries Patient}
& Hypertension; type 2 diabetes
& Prior-record diagnosis entries: \texttt{I10}, \texttt{E119} \\

4
& \textcolor{PatientText}{\sffamily\bfseries Patient}
& Metformin; lisinopril; aspirin
& Corresponding \texttt{medrecon.name} records \\

\bottomrule
\end{tabularx}
}

Following the event--turn lineage defined in
Section~\ref{sec:verification} and Equation~\eqref{eq:grounded-state},
the dialogue-visible state for this encounter evolves as follows.

\caseheading{\faListAlt\enspace Turn-wise accumulated grounded state}

Let \(\Delta Z_t^{\mathrm{C}}\) and \(\Delta Z_t^{\mathrm{P}}\)
denote source-supported facts first made explicit by the clinician and patient,
respectively, in Turn~\(t\). The dialogue-visible state accumulates as

\[
\Delta Z_t
=
\Delta Z_t^{\mathrm{C}}
\cup
\Delta Z_t^{\mathrm{P}},
\qquad
Z_0=\varnothing,
\qquad
Z_t=Z_{t-1}\cup\Delta Z_t .
\]

\begin{arrivalcontext}
\setlength{\jot}{3pt}
\[
\begin{aligned}
Z_1
&=
Z_0
\cup
\textcolor{PatientText}{
  \{\text{nausea, abdominal discomfort}\}_{\mathrm{P}}
},\\
Z_2
&=
Z_1
\cup
\textcolor{ClinicianText}{
  \{\text{heart rate }102\}_{\mathrm{C}}
}
\cup
\textcolor{PatientText}{
  \{\text{pain }6/10\}_{\mathrm{P}}
},\\
Z_3
&=
Z_2
\cup
\textcolor{PatientText}{
  \{\text{hypertension, type 2 diabetes}\}_{\mathrm{P}}
},\\
Z_4
&=
Z_3
\cup
\textcolor{PatientText}{
  \{\text{metformin, lisinopril, aspirin}\}_{\mathrm{P}}
}.
\end{aligned}
\]
\end{arrivalcontext}

{\scriptsize\sffamily\color{CasePurple!80}
Here, \(Z_t\) records what has become explicit in the conversation by
Turn~\(t\). A fact originating in the initial case record can therefore enter
\(Z_t\) when either role first says it, while only eligible patient information
receives acquisition credit in the benchmark metrics.}


\begin{closingstrip}

{\sffamily\bfseries\color{CasePurple}
\faClipboardList\enspace Patient-facing triage closing}

\smallskip

Thank you for walking me through that---I know you are uncomfortable.
Because your abdominal pain remains \(6/10\), you are still nauseated,
and your heart rate is elevated, I am assigning you
\textbf{ESI level 3}. A clinician will see you soon. If the pain suddenly
gets worse or you begin to feel faint while you are waiting, please tell
me immediately; we are here and paying attention.

\end{closingstrip}

{\scriptsize\sffamily\color{CasePurple!80}
\faLock\enspace
For this illustrative encounter, the assigned level matches the recorded
evaluation label \texttt{triage.acuity = 3}. The recorded label is not used to
generate the preceding questions or patient responses.}

\end{dialoguecase}


\definecolor{ComparePurple}{HTML}{514A82}
\definecolor{ComparePurpleLight}{HTML}{F2F0F8}
\definecolor{ComparePurpleBorder}{HTML}{C9C4DB}

\definecolor{CompareClinicianFill}{HTML}{EDF4FB}
\definecolor{CompareClinicianBorder}{HTML}{B8CDE3}
\definecolor{CompareClinicianText}{HTML}{315B7D}

\definecolor{ComparePatientFill}{HTML}{EAF6F1}
\definecolor{ComparePatientBorder}{HTML}{A9D2C3}
\definecolor{ComparePatientText}{HTML}{276A59}

\definecolor{CompareGray}{HTML}{606772}
\definecolor{CompareGrayLight}{HTML}{F3F4F6}
\definecolor{CompareBorder}{HTML}{D7DADE}
\definecolor{CompareRed}{HTML}{B34A4A}



\newtcolorbox{resourcecomparisonbox}{
    enhanced,
    colback=white,
    colframe=CompareBorder,
    boxrule=0.5pt,
    arc=1.2mm,
    left=7pt,
    right=7pt,
    top=7pt,
    bottom=7pt,
    before skip=0pt,
    after skip=0pt
}


\newtcolorbox{resourcepanel}[1]{
    enhanced,
    colback=white,
    colframe=CompareBorder,
    colbacktitle=ComparePurpleLight,
    coltitle=ComparePurple,
    title={#1},
    fonttitle=\sffamily\bfseries\small,
    boxrule=0.45pt,
    arc=1mm,
    left=5pt,
    right=5pt,
    top=5pt,
    bottom=5pt,
    toptitle=4pt,
    bottomtitle=4pt,
    before skip=0pt,
    after skip=0pt
}


\newtcolorbox{oursresourcepanel}[1]{
    enhanced,
    colback=white,
    colframe=CompareBorder,
    colbacktitle=ComparePurpleLight,
    coltitle=ComparePurple,
    title={#1},
    fonttitle=\sffamily\bfseries\small,
    boxrule=0.45pt,
    arc=1mm,
    left=6pt,
    right=6pt,
    top=6pt,
    bottom=6pt,
    toptitle=4pt,
    bottomtitle=4pt,
    before skip=0pt,
    after skip=0pt
}


\newtcolorbox{resourceinput}{
    enhanced,
    colback=CompareGrayLight,
    colframe=CompareBorder,
    boxrule=0.35pt,
    arc=0.8mm,
    left=4pt,
    right=4pt,
    top=4pt,
    bottom=4pt,
    before skip=2pt,
    after skip=5pt,
    fontupper=\scriptsize
}


\newtcolorbox{resourcepurpleinput}{
    enhanced,
    colback=ComparePurpleLight,
    colframe=ComparePurpleBorder,
    boxrule=0.35pt,
    arc=0.8mm,
    left=4pt,
    right=4pt,
    top=4pt,
    bottom=4pt,
    before skip=2pt,
    after skip=5pt,
    fontupper=\scriptsize
}


\newtcolorbox{resourceoutput}{
    enhanced,
    colback=CompareGrayLight,
    colframe=CompareBorder,
    boxrule=0.35pt,
    arc=0.8mm,
    left=4pt,
    right=4pt,
    top=4pt,
    bottom=4pt,
    before skip=4pt,
    after skip=4pt,
    fontupper=\scriptsize
}


\newtcolorbox{resourcepurpleoutput}{
    enhanced,
    colback=ComparePurpleLight,
    colframe=ComparePurpleBorder,
    boxrule=0.35pt,
    arc=0.8mm,
    left=4pt,
    right=4pt,
    top=4pt,
    bottom=4pt,
    before skip=4pt,
    after skip=4pt,
    fontupper=\scriptsize
}


\newtcolorbox{resourceTurnBox}{
    enhanced,
    colback=white,
    colframe=ComparePurpleBorder,
    boxrule=0.40pt,
    arc=0.8mm,
    left=5pt,
    right=5pt,
    top=5pt,
    bottom=5pt,
    before skip=0pt,
    after skip=0pt
}


\newcommand{\resourceclinician}[1]{%
    \begin{tcolorbox}[
        enhanced,
        colback=CompareClinicianFill,
        colframe=CompareClinicianBorder,
        boxrule=0.30pt,
        arc=0.8mm,
        left=4pt,
        right=4pt,
        top=3pt,
        bottom=3pt,
        before skip=3pt,
        after skip=3pt
    ]
    {\scriptsize
    \textcolor{CompareClinicianText}{\textbf{Clinician:}} #1}
    \end{tcolorbox}
}

\newcommand{\resourcepatient}[1]{%
    \begin{tcolorbox}[
        enhanced,
        colback=ComparePatientFill,
        colframe=ComparePatientBorder,
        boxrule=0.30pt,
        arc=0.8mm,
        left=4pt,
        right=4pt,
        top=3pt,
        bottom=3pt,
        before skip=3pt,
        after skip=3pt
    ]
    {\scriptsize
    \textcolor{ComparePatientText}{\textbf{Patient:}} #1}
    \end{tcolorbox}
}


\newcommand{\resourcestate}[1]{%
    \begin{tcolorbox}[
        enhanced,
        colback=ComparePurpleLight,
        colframe=ComparePurpleBorder,
        boxrule=0.30pt,
        arc=0.7mm,
        left=3pt,
        right=3pt,
        top=3pt,
        bottom=3pt,
        before skip=4pt,
        after skip=0pt
    ]
    \centering
    {\scriptsize\sffamily\bfseries
    \textcolor{ComparePurple}{#1}}
    \end{tcolorbox}
}


\newcommand{\resourcecapline}[1]{%
    \par\smallskip
    \begin{center}
        \scriptsize\sffamily #1
    \end{center}
}

\newcommand{\resourcecapno}[1]{%
    \textcolor{CompareRed}{%
        \(\times\)\enspace #1%
    }%
}

\newcommand{\resourcecapyes}[1]{%
    \textcolor{ComparePurple}{%
        \(\checkmark\)\enspace #1%
    }%
}


\begin{figure*}[t]
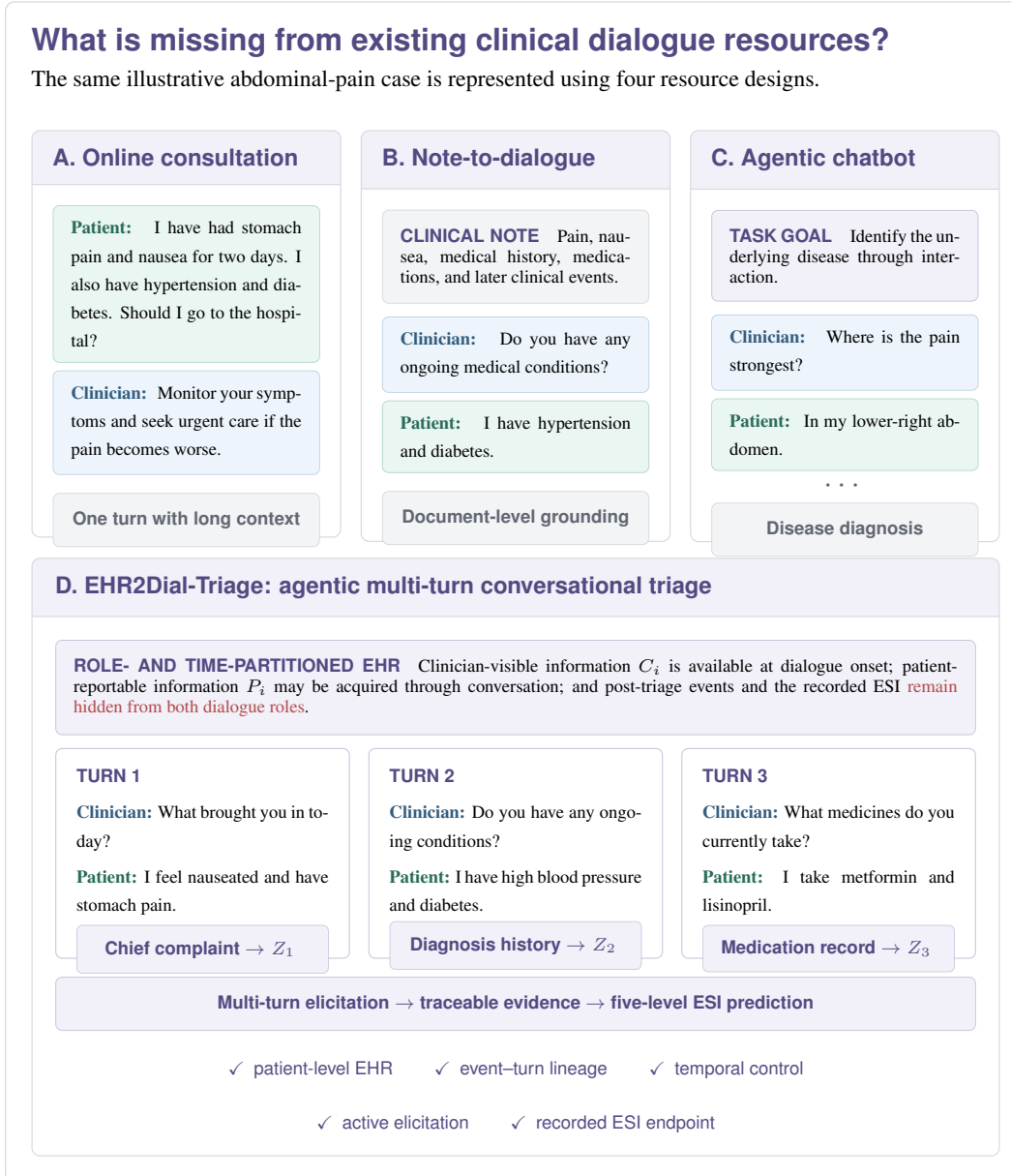

\centering

\begin{resourcecomparisonbox}

{\sffamily\bfseries\color{ComparePurple}\large
What is missing from existing clinical dialogue resources?}

\smallskip

{\small
The same illustrative abdominal-pain case is represented using four
resource designs.}

\medskip


\begin{minipage}[t]{0.32\linewidth}
\vspace{0pt}

\begin{resourcepanel}{A. Online consultation}

\begin{minipage}[t][4.25cm][t]{\linewidth}

\resourcepatient{
I have had stomach pain and nausea for two days. I also have hypertension
and diabetes. Should I go to the hospital?
}

\resourceclinician{
Monitor your symptoms and seek urgent care if the pain becomes worse.
}

\vfill

\begin{resourceoutput}
\centering
{\sffamily\bfseries\color{CompareGray}
One turn with long context}
\end{resourceoutput}

\resourcecapline{
\resourcecapno{No interaction}
}

\end{minipage}

\end{resourcepanel}

\end{minipage}
\hfill
%
%
\begin{minipage}[t]{0.32\linewidth}
\vspace{0pt}

\begin{resourcepanel}{B. Note-to-dialogue}

\begin{minipage}[t][4.25cm][t]{\linewidth}

\begin{resourceinput}
{\sffamily\bfseries\color{ComparePurple}CLINICAL NOTE}\quad
Pain, nausea, medical history, medications, and later clinical events.
\end{resourceinput}

\resourceclinician{
Do you have any ongoing medical conditions?
}

\resourcepatient{
I have hypertension and diabetes.
}

\vfill

\begin{resourceoutput}
\centering
{\sffamily\bfseries\color{CompareGray}
Document-level grounding}
\end{resourceoutput}

\resourcecapline{
\resourcecapno{No event and time tracing}
}

\end{minipage}

\end{resourcepanel}

\end{minipage}
\hfill
%
%
\begin{minipage}[t]{0.32\linewidth}
\vspace{0pt}

\begin{resourcepanel}{C. Agentic chatbot}

\begin{minipage}[t][4.25cm][t]{\linewidth}

\begin{resourcepurpleinput}
{\sffamily\bfseries\color{ComparePurple}TASK GOAL}\quad
Identify the underlying disease through interaction.
\end{resourcepurpleinput}

\resourceclinician{
Where is the pain strongest?
}

\resourcepatient{
In my lower-right abdomen.
}

\begin{center}
\vspace{-2pt}
{\color{CompareGray}\large\(\cdots\)}
\vspace{-5pt}
\end{center}

\vfill

\begin{resourceoutput}
\centering
{\sffamily\bfseries\color{CompareGray}
Disease diagnosis}
\end{resourceoutput}

\resourcecapline{
\resourcecapno{Not designed for triage}
}

\end{minipage}

\end{resourcepanel}

\end{minipage}

\medskip


\begin{oursresourcepanel}{
D. EHR2Dial-Triage: agentic multi-turn conversational triage
}

\begin{resourcepurpleinput}

{\sffamily\bfseries\color{ComparePurple}
ROLE- AND TIME-PARTITIONED EHR}\quad
Clinician-visible information \(C_i\) is available at dialogue onset;
patient-reportable information \(P_i\) may be acquired through conversation;
and post-triage events and the recorded ESI
\textcolor{CompareRed}{remain hidden from both dialogue roles}.

\end{resourcepurpleinput}


\begin{minipage}[t]{0.32\linewidth}
\vspace{0pt}

\begin{resourceTurnBox}

\begin{minipage}[t][2.30cm][t]{\linewidth}

{\scriptsize\sffamily\bfseries\color{ComparePurple}
TURN 1}

\smallskip

{\scriptsize
\textcolor{CompareClinicianText}{\textbf{Clinician:}}
What brought you in today?}

\smallskip

{\scriptsize
\textcolor{ComparePatientText}{\textbf{Patient:}}
I feel nauseated and have stomach pain.}

\vfill

\resourcestate{
Chief complaint \(\rightarrow Z_1\)
}

\end{minipage}

\end{resourceTurnBox}

\end{minipage}
\hfill
%
%
\begin{minipage}[t]{0.32\linewidth}
\vspace{0pt}

\begin{resourceTurnBox}

\begin{minipage}[t][2.30cm][t]{\linewidth}

{\scriptsize\sffamily\bfseries\color{ComparePurple}
TURN 2}

\smallskip

{\scriptsize
\textcolor{CompareClinicianText}{\textbf{Clinician:}}
Do you have any ongoing conditions?}

\smallskip

{\scriptsize
\textcolor{ComparePatientText}{\textbf{Patient:}}
I have high blood pressure and diabetes.}

\vfill

\resourcestate{
Diagnosis history \(\rightarrow Z_2\)
}

\end{minipage}

\end{resourceTurnBox}

\end{minipage}
\hfill
%
%
\begin{minipage}[t]{0.32\linewidth}
\vspace{0pt}

\begin{resourceTurnBox}

\begin{minipage}[t][2.30cm][t]{\linewidth}

{\scriptsize\sffamily\bfseries\color{ComparePurple}
TURN 3}

\smallskip

{\scriptsize
\textcolor{CompareClinicianText}{\textbf{Clinician:}}
What medicines do you currently take?}

\smallskip

{\scriptsize
\textcolor{ComparePatientText}{\textbf{Patient:}}
I take metformin and lisinopril.}

\vfill

\resourcestate{
Medication record \(\rightarrow Z_3\)
}

\end{minipage}

\end{resourceTurnBox}

\end{minipage}

\smallskip


\begin{resourcepurpleoutput}

\centering

{\sffamily\bfseries\color{ComparePurple}
Multi-turn elicitation
\(\rightarrow\)
traceable evidence
\(\rightarrow\)
five-level ESI prediction}

\end{resourcepurpleoutput}


\resourcecapline{
\resourcecapyes{patient-level EHR}
\qquad
\resourcecapyes{event--turn lineage}
\qquad
\resourcecapyes{temporal control}
}

\resourcecapline{
\resourcecapyes{active elicitation}
\qquad
\resourcecapyes{recorded ESI endpoint}
}

\end{oursresourcepanel}

\end{resourcecomparisonbox}

\caption{\textbf{Illustrative comparison of clinical dialogue resource
designs.}
Online consultations commonly contain few long turns; note-to-dialogue
resources may provide document-level rather than event-level grounding; and
representative agentic clinical benchmarks often target diagnosis.
EHR2Dial-Triage combines multi-turn information elicitation, patient-level
EHR grounding, event--turn lineage, temporal information control, and
recorded five-level ESI prediction. All utterances are schematic and use the
same illustrative case.}

\label{fig:resource-design-comparison}

\end{figure*}
\section{Frozen Benchmark-Cohort Statistics}
\label{app:benchmark-cohort-statistics}

\setcounter{dbltopnumber}{2}
\renewcommand{\dbltopfraction}{0.92}
\renewcommand{\dblfloatpagefraction}{0.85}
\renewcommand{\textfraction}{0.05}
\setlength{\dbltextfloatsep}{8pt plus 2pt minus 2pt}

\subsection{Population and Label Composition}
\label{app:benchmark-population}

The figures in this section describe the frozen 4,041-encounter benchmark
cohort used for the matched corpus analysis. These encounters correspond to
3,888 unique patients, and all 4,041 have a recorded five-level ESI label. The
cohort is a deliberately selected, acuity-enriched subset of MIMIC-IV-ED v2.2;
its composition therefore characterizes the benchmark cases rather than
emergency-department prevalence.

\begin{figure*}[!t]
    \centering
    \includegraphics[
        width=0.80\textwidth,
        keepaspectratio
    ]{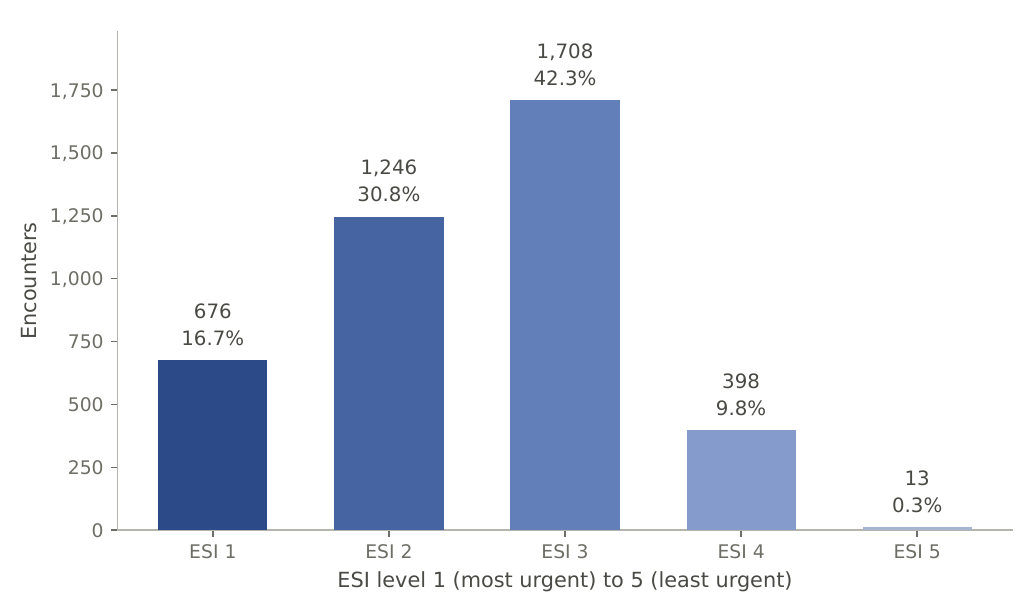}
    \caption{\textbf{Recorded triage acuity in the frozen benchmark cohort.}
    Bars show encounter counts and percentages of all 4,041 cohort encounters
    for ESI levels 1--5. All cohort encounters have a recorded ESI value.}
    \label{fig:benchmark-cohort-esi}
\end{figure*}

\begin{figure*}[!t]
    \centering
    \begin{minipage}[t]{0.485\textwidth}
        \centering
        \includegraphics[
            height=0.135\textheight,
            trim=15bp 370bp 10bp 15bp,
            clip
        ]{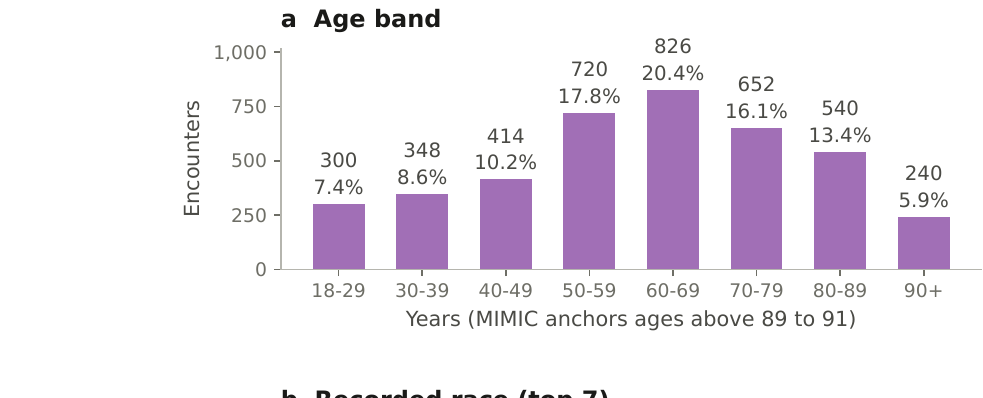}
    \end{minipage}\hfill
    \begin{minipage}[t]{0.485\textwidth}
        \centering
        \includegraphics[
            height=0.135\textheight,
            trim=5bp 180bp 5bp 180bp,
            clip
        ]{singles/fig_benchmark_cohort_04_demographics.pdf}
    \end{minipage}
    \par\vspace{1mm}
    \begin{minipage}[t]{0.485\textwidth}
        \centering
        \textbf{c\enspace Recorded sex}\par\vspace{1pt}
        \includegraphics[
            height=0.11\textheight,
            trim=15bp 8bp 215bp 430bp,
            clip
        ]{singles/fig_benchmark_cohort_04_demographics.pdf}
    \end{minipage}\hfill
    \begin{minipage}[t]{0.485\textwidth}
        \centering
        \textbf{d\enspace Arrival mode}\par\vspace{1pt}
        \includegraphics[
            height=0.11\textheight,
            trim=280bp 8bp 2bp 430bp,
            clip
        ]{singles/fig_benchmark_cohort_04_demographics.pdf}
    \end{minipage}
    \caption{\textbf{Population and arrival characteristics in the frozen
    benchmark cohort.} (a) Age band. (b) The seven most frequent recorded race
    categories. (c) Recorded sex. (d) Arrival mode. Percentages use all 4,041
    cohort encounters as the denominator.}
    \label{fig:benchmark-cohort-demographics}
\end{figure*}

Figure~\ref{fig:benchmark-cohort-esi} shows the cohort's recorded acuity
distribution. ESI-1 accounts for 16.73\% of encounters, ESI-2 for 30.83\%,
ESI-3 for 42.27\%, ESI-4 for 9.85\%, and ESI-5 for 0.32\%.
Figure~\ref{fig:benchmark-cohort-demographics} reports the corresponding age,
race, sex, and arrival-mode composition. Categories and missing values retain
their source representation.

\begin{table*}[h]
\centering
\caption{\textbf{Source fields and their use in benchmark construction.}
Clinician-visible information \(C_i\) is available to the clinician from the
beginning of the interaction, whereas patient-reportable information \(P_i\)
must be elicited through dialogue. A fact may belong to both partitions when
it is documented in the pre-arrival record and can also be plausibly reported
by the patient. Future-hidden and evaluation-only fields are retained in the
episode but withheld from both dialogue roles. The verifier uses future-hidden
fields to detect post-triage leakage, while evaluation-only fields are used
only for labels or case characterization. Recorded acuity is retained solely
as the prediction target \(Y_i\).}
\label{tab:source-fields}

\scriptsize
\setlength{\tabcolsep}{4pt}
\renewcommand{\arraystretch}{1.12}

\begin{tabularx}{\textwidth}{
    @{}
    >{\raggedright\arraybackslash}p{2.9cm}
    >{\raggedright\arraybackslash}X
    >{\centering\arraybackslash}p{2.0cm}
    >{\raggedright\arraybackslash}X
    @{}
}
\toprule
Source & Representative content & Benchmark use & Rationale \\
\midrule

\texttt{triage}
& Initial vital signs
& \(C_i\)
& Available to the clinician in the initial triage record \\

\texttt{triage}
& Chief complaint and initial pain information
& \(P_i\)
& Patient-reportable information that may be elicited in patient language \\

\texttt{triage}
& Recorded acuity
& \(Y_i\) only
& Downstream five-level ESI prediction target \\

\texttt{edstays}
& Arrival time and encounter identifiers
& Case anchor
& Defines the encounter time \(\tau_i\) and links source records \\

\texttt{edstays}
& Age and sex
& \(C_i,\;P_i\)
& Available as structured encounter information before the dialogue begins \\

\texttt{edstays}
& Arrival mode
& \(P_i\)
& Supported arrival context that the patient can report during dialogue \\

\texttt{edstays}
& Disposition
& Evaluation-only
& Determined after triage and withheld from both dialogue roles \\

\texttt{medrecon}
& Reconciled home medications
& \(P_i\)
& Pre-arrival medication history that may be reported by the patient \\

\texttt{diagnoses\_icd\_prior}
& Diagnoses recorded during earlier encounters
& \(C_i,\;P_i\)
& Pre-arrival history that may be visible in the record and plausibly
reported by the patient \\

\texttt{vitalsign}
& Measurements recorded after \(\tau_i\)
& Future-hidden
& Unavailable at triage onset and retained only for temporal-leakage detection \\

\texttt{pyxis}
& ED medication-dispensing events
& Future-hidden
& Reflect post-arrival care and are retained only for leakage detection \\

\texttt{diagnosis}
& Discharge diagnoses
& Evaluation-only
& Post-triage information used for case characterization, never as model input \\

\bottomrule
\end{tabularx}
\end{table*}

\subsection{Triage Measurements and Record Completeness}
\label{app:benchmark-measurements}

\begin{figure*}[!t]
    \centering
    \begin{minipage}[t]{0.318\textwidth}
        \includegraphics[width=\linewidth,trim=20bp 352bp 250bp 20bp,clip]
        {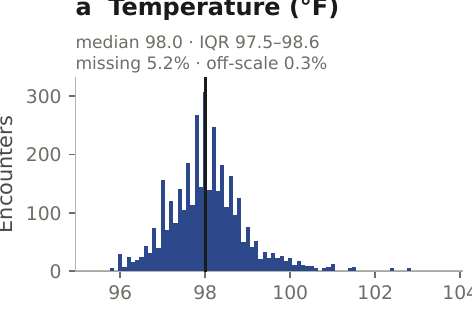}
    \end{minipage}\hfill
    \begin{minipage}[t]{0.318\textwidth}
        \includegraphics[width=\linewidth,trim=250bp 352bp 20bp 20bp,clip]
        {singles/fig_benchmark_cohort_02_vital_signs.pdf}
    \end{minipage}\hfill
    \begin{minipage}[t]{0.318\textwidth}
        \includegraphics[width=\linewidth,trim=20bp 155bp 250bp 165bp,clip]
        {singles/fig_benchmark_cohort_02_vital_signs.pdf}
    \end{minipage}
    \par\vspace{1mm}
    \begin{minipage}[t]{0.318\textwidth}
        \includegraphics[width=\linewidth,trim=250bp 155bp 20bp 165bp,clip]
        {singles/fig_benchmark_cohort_02_vital_signs.pdf}
    \end{minipage}\hfill
    \begin{minipage}[t]{0.318\textwidth}
        \includegraphics[width=\linewidth,trim=20bp 5bp 250bp 330bp,clip]
        {singles/fig_benchmark_cohort_02_vital_signs.pdf}
    \end{minipage}\hfill
    \begin{minipage}[t]{0.318\textwidth}
        \includegraphics[width=\linewidth,trim=250bp 5bp 20bp 330bp,clip]
        {singles/fig_benchmark_cohort_02_vital_signs.pdf}
    \end{minipage}
    \caption{\textbf{Distributions of first-recorded triage vital signs in the
    frozen benchmark cohort.} (a) Temperature. (b) Heart rate.
    (c) Respiratory rate. (d) Oxygen saturation. (e) Systolic blood pressure.
    (f) Diastolic blood pressure. Panel annotations report the median, IQR,
    missingness, and off-scale share.}
    \label{fig:benchmark-cohort-vitals}
\end{figure*}

\begin{figure*}[!t]
    \centering
    \begin{minipage}[t]{0.485\textwidth}
        \includegraphics[width=\linewidth,trim=5bp 160bp 5bp 10bp,clip]
        {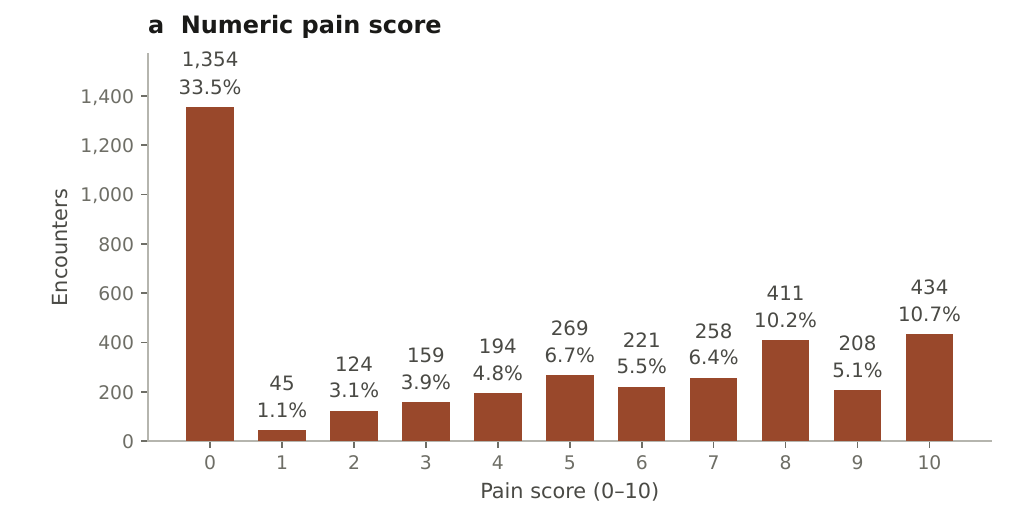}
    \end{minipage}\hfill
    \begin{minipage}[t]{0.485\textwidth}
        \includegraphics[width=\linewidth,trim=5bp 0bp 5bp 270bp,clip]
        {singles/fig_benchmark_cohort_03_pain_score.pdf}
    \end{minipage}
    \caption{\textbf{Self-reported pain values at triage in the frozen
    benchmark cohort.} (a) Distribution of valid integer scores from 0 to 10.
    (b) Counts of non-integer, out-of-range, free-text, and missing entries.}
    \label{fig:benchmark-cohort-pain}
\end{figure*}

Figures~\ref{fig:benchmark-cohort-vitals} and
\ref{fig:benchmark-cohort-pain} summarize the first recorded triage measurements
for each ED stay. Vital-sign distributions are shown together with their
median, interquartile range (IQR), missingness, and off-scale share. Pain is
handled separately because the source field includes both numeric scores and
nonstandard text entries; these entries are displayed rather than coerced to a
numeric value.

\begin{figure*}[!t]
    \centering
    \begin{minipage}[t]{0.318\textwidth}
        \includegraphics[width=\linewidth,trim=5bp 340bp 245bp 20bp,clip]
        {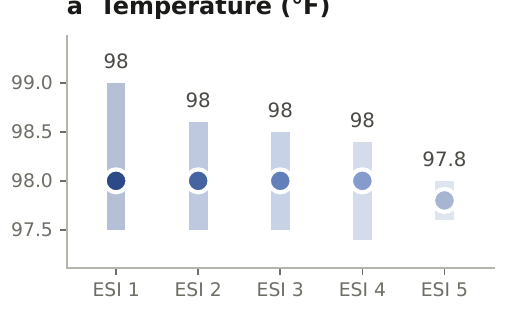}
    \end{minipage}\hfill
    \begin{minipage}[t]{0.318\textwidth}
        \includegraphics[width=\linewidth,trim=245bp 340bp 5bp 20bp,clip]
        {singles/fig_benchmark_cohort_06_vitals_by_esi.pdf}
    \end{minipage}\hfill
    \begin{minipage}[t]{0.318\textwidth}
        \includegraphics[width=\linewidth,trim=5bp 174bp 245bp 174bp,clip]
        {singles/fig_benchmark_cohort_06_vitals_by_esi.pdf}
    \end{minipage}
    \par\vspace{1mm}
    \begin{minipage}[t]{0.318\textwidth}
        \includegraphics[width=\linewidth,trim=245bp 174bp 5bp 174bp,clip]
        {singles/fig_benchmark_cohort_06_vitals_by_esi.pdf}
    \end{minipage}\hfill
    \begin{minipage}[t]{0.318\textwidth}
        \includegraphics[width=\linewidth,trim=5bp 8bp 245bp 332bp,clip]
        {singles/fig_benchmark_cohort_06_vitals_by_esi.pdf}
    \end{minipage}\hfill
    \begin{minipage}[t]{0.318\textwidth}
        \includegraphics[width=\linewidth,trim=245bp 8bp 5bp 332bp,clip]
        {singles/fig_benchmark_cohort_06_vitals_by_esi.pdf}
    \end{minipage}
    \caption{\textbf{Triage vital signs by recorded ESI level.}
    (a) Temperature. (b) Heart rate. (c) Respiratory rate. (d) Oxygen
    saturation. (e) Systolic blood pressure. (f) Diastolic blood pressure.
    Points and intervals show the median and IQR within each ESI level.}
    \label{fig:benchmark-cohort-vitals-by-esi}
\end{figure*}

Figure~\ref{fig:benchmark-cohort-vitals-by-esi} stratifies the same vital signs by
recorded ESI. The point-and-interval encoding is consistent across all six
panels: points denote medians and vertical intervals denote IQRs. These are
descriptive associations and should not be interpreted as isolated determinants
of acuity. The missingness annotations in
Figures~\ref{fig:benchmark-cohort-vitals} and
\ref{fig:benchmark-cohort-pain} complete the measurement audit; every displayed
core triage field is recorded for at least 94.8\% of cohort encounters.

\subsection{Arrival, Flow, and Acuity Heterogeneity}
\label{app:benchmark-arrival-acuity}

\begin{figure*}[!t]
    \centering
    \includegraphics[
        width=0.90\textwidth,
        keepaspectratio
    ]{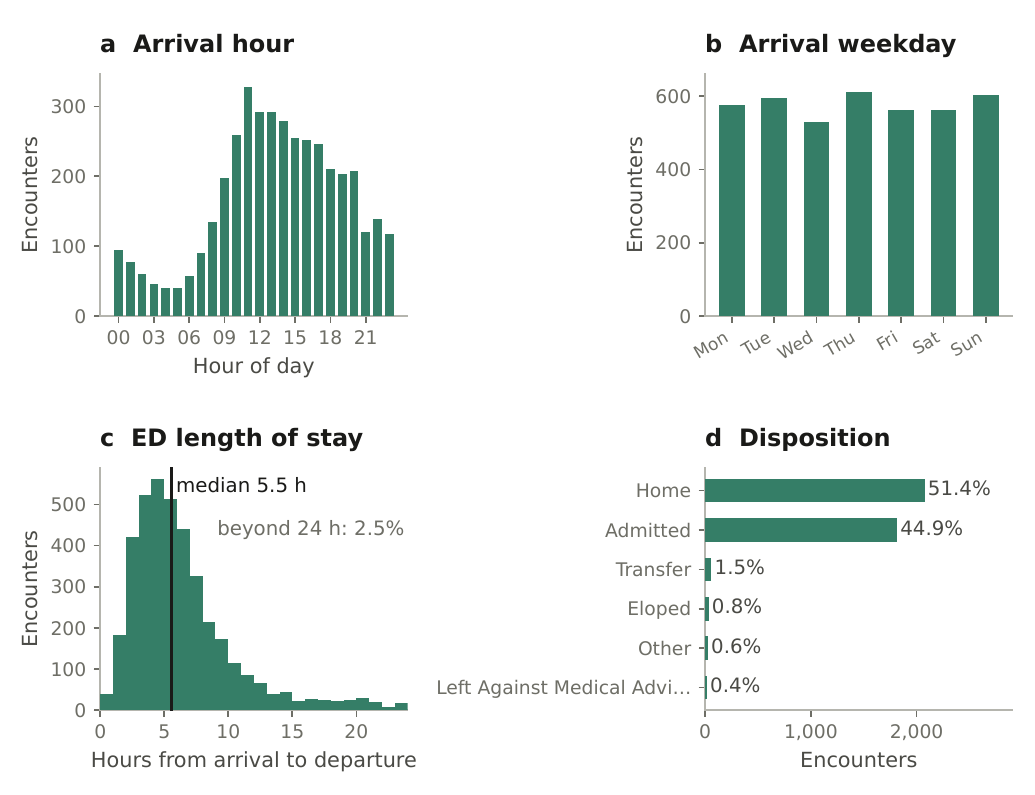}
    \caption{\textbf{Arrival timing and subsequent ED flow in the frozen
    benchmark cohort.} (a) Arrival hour. (b) Arrival weekday. (c) ED length of
    stay, truncated at 24 hours for display with the excluded share annotated.
    (d) Recorded disposition.}
    \label{fig:benchmark-cohort-arrival-flow}
\end{figure*}

\begin{figure*}[!t]
    \centering
    \begin{minipage}[t]{0.485\textwidth}
        \includegraphics[width=\linewidth,trim=0bp 300bp 0bp 18bp,clip]
        {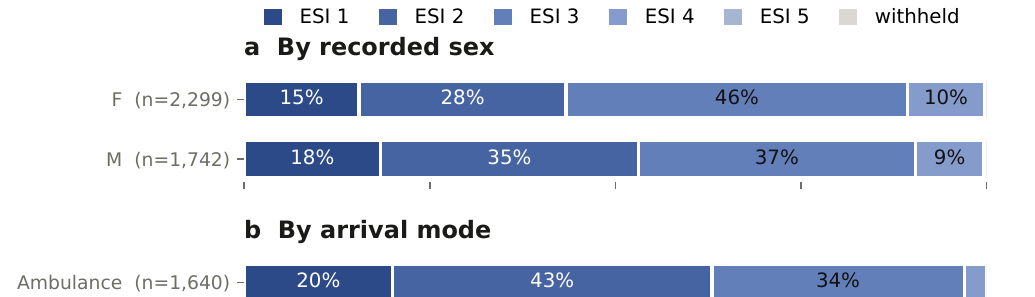}
    \end{minipage}\hfill
    \begin{minipage}[t]{0.485\textwidth}
        \includegraphics[width=\linewidth,trim=0bp 180bp 0bp 140bp,clip]
        {singles/fig_benchmark_cohort_07_acuity_by_subgroup.pdf}
    \end{minipage}
    \par\vspace{1mm}
    \includegraphics[
        width=0.82\textwidth,
        trim=0bp 8bp 0bp 205bp,
        clip
    ]{singles/fig_benchmark_cohort_07_acuity_by_subgroup.pdf}
    \caption{\textbf{Recorded acuity mix across benchmark-cohort subgroups.}
    Row-normalized ESI distributions are shown (a) by recorded sex, (b) by
    arrival mode, and (c) by age band. Labels are omitted for very small
    segments to maintain legibility.}
    \label{fig:benchmark-cohort-acuity-subgroups}
\end{figure*}

The source data also capture arrival timing, ED length of stay, and disposition.
Figure~\ref{fig:benchmark-cohort-arrival-flow} summarizes these process variables;
date shifting preserves time of day and weekday, allowing their distributions
to be reported. The median cohort ED stay is 5.5 hours. These post-triage process
variables are descriptive only and remain excluded from both dialogue roles and
the benchmark reader. Figure~\ref{fig:benchmark-cohort-acuity-subgroups} shows how the
recorded acuity mix varies across sex, arrival mode, and age band. Each bar is
normalized within its subgroup, so the figure compares composition rather than
subgroup size. These summaries characterize cohort heterogeneity and do not
define separate benchmark evaluation strata.

\begin{figure*}[!t]
    \centering
    \begin{minipage}[t]{0.485\textwidth}
        \includegraphics[width=\linewidth,trim=0bp 215bp 0bp 8bp,clip]
        {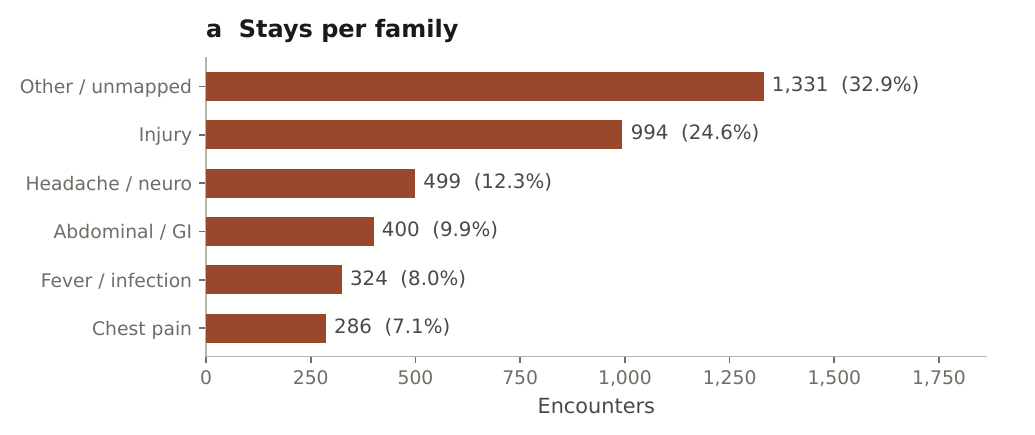}
    \end{minipage}\hfill
    \begin{minipage}[t]{0.485\textwidth}
        \includegraphics[width=\linewidth,trim=0bp 2bp 0bp 185bp,clip]
        {singles/fig_benchmark_cohort_08_presentation_families.pdf}
    \end{minipage}
    \caption{\textbf{Presentation families and recorded acuity in the
    frozen benchmark cohort.} (a) Counts and shares of rule-based
    chief-complaint families. (b) Row-normalized ESI distribution within each
    family. Labels are omitted for very small segments.}
    \label{fig:benchmark-cohort-presentations}
\end{figure*}

Finally, Figure~\ref{fig:benchmark-cohort-presentations} groups chief-complaint
strings into broad rule-based presentation families. The upper panel reports
the number and share of stays in each family; the lower panel reports the
row-normalized ESI composition within those same families. The substantial
within-family variation reinforces that a complaint category does not uniquely
determine encounter acuity.

These statistics directly describe the frozen benchmark cohort analyzed here.
Because the cohort is acuity enriched by construction, neither its label mix
nor performance measured on it should be interpreted as an estimate of
real-world ESI prevalence or deployment performance.


\section{Evaluation Metric Definitions}
\label{app:evaluation-details}

This section provides the complete definitions of the evaluation metrics used
in Experiments~I and II and in the safety analysis. Unless otherwise stated,
metrics are first computed at the encounter level and then macro-averaged over
the \(N=1{,}013\) evaluation encounters. The recorded ESI label is denoted by
\(y_i\in\{1,2,3,4,5\}\), where a smaller value indicates higher urgency.

\subsection{Clinician Questioning and Information Acquisition}
\label{app:elicitation-metrics}

For encounter \(i\), let \(B_i\) denote the set of eligible patient-reportable
facts available to the patient agent. These facts are derived from
temporally eligible EHR events and exclude clinician-visible arrival
information, the recorded ESI label, and post-triage information. Let
\(R_i^{(r)}\subseteq B_i\) be the set of facts newly disclosed in the accepted
patient response at round \(r\), as recorded by
\texttt{new\_fact\_ids}. The cumulative set of acquired facts by Turn \(t\) is

\begin{equation}
S_i^{(t)}
=
\bigcup_{r=1}^{t} R_i^{(r)}.
\label{eq:cumulative-acquired-facts}
\end{equation}

A fact is counted once at its first accepted appearance. Repeated mentions do
not increase coverage. Facts mentioned in rejected generations or unsupported
claims are not included in \(S_i^{(t)}\).

\paragraph{Question-target distribution.}
Each substantive clinician question is assigned one primary information target
using the same fixed target annotation procedure for all models. Let
\(z_{i,r}\in\mathcal{C}\) be the assigned target of the clinician question at
round \(r\), where \(\mathcal{C}\) is the set of target categories shown in
Figure~\ref{fig:question-targets}. For clinician model \(m\), the share assigned
to target \(c\) is

\begin{equation}
\operatorname{TargetShare}_{m,c}
=
\frac{
\sum_{i}\sum_{r}
\mathbf{1}\!\left[z_{i,r}=c\right]
}{
\sum_{i}\sum_{r}
\mathbf{1}\!\left[z_{i,r}\in\mathcal{C}\right]
}.
\label{eq:question-target-share}
\end{equation}

The shares are calculated over all valid clinician questions generated by the
model and sum to one across target categories.

\paragraph{Question yield.}
Question yield measures the proportion of clinician questions followed by at
least one newly disclosed, source-supported patient fact. For clinician model
\(m\), it is defined as

\begin{equation}
\operatorname{QYield}_{m}
=
\frac{
\sum_{i}\sum_{r}
\mathbf{1}\!\left[|R_i^{(r)}|>0\right]
}{
\sum_{i} T_i
},
\label{eq:question-yield}
\end{equation}

where \(T_i\) is the number of completed clinician--patient rounds for
encounter \(i\). A response containing only previously disclosed facts has
zero yield for that round.

\paragraph{Cumulative fact coverage.}
\textsc{FactCov@}\(t\) measures the proportion of eligible patient facts
acquired by checkpoint \(t\). We first calculate coverage within each
encounter and then average across encounters:

\begin{equation}
\operatorname{FactCov@}t
=
\frac{1}{N}
\sum_{i=1}^{N}
\frac{|S_i^{(t)}|}{|B_i|}.
\label{eq:fact-coverage}
\end{equation}

This encounter-level macro-average gives equal weight to each encounter,
independent of the size of its fact bank. Values reported in
Table~\ref{tab:exp1-complete-main} are multiplied by \(100\) and presented as
percentages.

\paragraph{History and home-medication coverage.}
Let \(B_i^{\mathrm{HM}}\subseteq B_i\) contain the eligible medical-history and
home-medication facts for encounter \(i\). Cumulative coverage of this subset
is

\begin{equation}
\operatorname{HistoryMedCov@}t
=
\frac{1}{|\mathcal{I}_{\mathrm{HM}}|}
\sum_{i\in\mathcal{I}_{\mathrm{HM}}}
\frac{
|S_i^{(t)}\cap B_i^{\mathrm{HM}}|
}{
|B_i^{\mathrm{HM}}|
},
\label{eq:history-med-coverage}
\end{equation}

where

\begin{equation}
\mathcal{I}_{\mathrm{HM}}
=
\left\{
i:|B_i^{\mathrm{HM}}|>0
\right\}.
\end{equation}

Encounters without an eligible history or home-medication fact are excluded
from this metric because their within-encounter recall denominator is zero.
As with \textsc{FactCov}, repeated disclosures receive no additional credit.

\paragraph{Frozen-reader entropy.}
We use a fixed patient-disjoint multinomial logistic regression reader to
measure how its ESI distribution changes as source-supported information
accumulates. The reader is trained once and remains unchanged across clinician
models and evaluation checkpoints.

Let \(\mathbf{x}_i^{(t)}\) be the reader representation constructed from the
clinician-visible arrival context \(D_i\) and the cumulative acquired fact set
\(S_i^{(t)}\). The reader produces the five-class distribution

\begin{equation}
\mathbf{p}_i^{(t)}
=
f_{\mathrm{reader}}\!
\left(\mathbf{x}_i^{(t)}\right),
\qquad
\sum_{k=1}^{5}p_{i,k}^{(t)}=1.
\end{equation}

The predictive entropy for encounter \(i\) is

\begin{equation}
H_i^{(t)}
=
-\sum_{k=1}^{5}
p_{i,k}^{(t)}
\log p_{i,k}^{(t)}.
\label{eq:reader-entropy-instance}
\end{equation}

Mean reader entropy at checkpoint \(t\) is

\begin{equation}
\operatorname{Entropy@}t
=
\frac{1}{N}
\sum_{i=1}^{N}
H_i^{(t)}.
\label{eq:reader-entropy}
\end{equation}

Natural logarithms are used, so entropy is reported in nats. Lower entropy
indicates a more concentrated predictive distribution. Entropy measures
distributional concentration and does not determine whether the most probable
ESI level matches the recorded decision.

When entropy reduction is reported, it is calculated relative to the
arrival-context reader input:

\begin{equation}
\Delta H@t
=
\frac{1}{N}
\sum_{i=1}^{N}
\left(
H_i^{(0)}-H_i^{(t)}
\right),
\label{eq:entropy-reduction}
\end{equation}

where \(H_i^{(0)}\) is obtained from \(D_i\) before any patient facts have been
acquired.

\subsection{Dialogue-Based ESI Prediction}
\label{app:esi-prediction-metrics}

For each encounter, the evaluated model returns a five-class score vector
\(\mathbf{p}_i=(p_{i,1},\ldots,p_{i,5})\) under the common output protocol.
The predicted ESI level is

\begin{equation}
\hat{y}_i
=
\arg\max_{k\in\{1,\ldots,5\}}p_{i,k}.
\label{eq:predicted-esi}
\end{equation}

The same parsed score vectors are used for Macro-AUC and confidence analysis,
and the discrete predictions are used for Macro-F1, QWK, and under-triage
analysis.

\paragraph{Macro-AUC.}
For each ESI level \(k\), we calculate a one-vs-rest area under the receiver
operating characteristic curve using \(p_{i,k}\) as the class score.
Macro-AUC is the unweighted mean across the five levels:

\begin{equation}
\operatorname{MacroAUC}
=
\frac{1}{5}
\sum_{k=1}^{5}
\operatorname{AUC}_{k}.
\label{eq:macro-auc}
\end{equation}

This metric gives equal weight to each ESI level and evaluates discrimination
using the full class-score vector.

\paragraph{Macro-F1.}
For ESI level \(k\), precision, recall, and F1 are

\begin{align}
P_k
&=
\frac{\mathrm{TP}_k}
{\mathrm{TP}_k+\mathrm{FP}_k},
\\
R_k
&=
\frac{\mathrm{TP}_k}
{\mathrm{TP}_k+\mathrm{FN}_k},
\\
F1_k
&=
\frac{2P_kR_k}{P_k+R_k}.
\end{align}

Macro-F1 is

\begin{equation}
\operatorname{MacroF1}
=
\frac{1}{5}
\sum_{k=1}^{5}F1_k.
\label{eq:macro-f1}
\end{equation}

A class-level F1 score is set to zero when its precision and recall are both
undefined because the model produces no positive prediction for that class.

\paragraph{Quadratic weighted kappa.}
Quadratic weighted kappa evaluates agreement while accounting for the ordinal
distance between the predicted and recorded ESI levels. Let \(O_{ab}\) be the
normalized observed confusion matrix and \(E_{ab}\) the expected matrix
obtained from the corresponding prediction and label marginals. The quadratic
weight for cell \((a,b)\) is

\begin{equation}
w_{ab}
=
\frac{(a-b)^2}{(5-1)^2}.
\end{equation}

QWK is then

\begin{equation}
\operatorname{QWK}
=
1-
\frac{
\sum_{a=1}^{5}\sum_{b=1}^{5}w_{ab}O_{ab}
}{
\sum_{a=1}^{5}\sum_{b=1}^{5}w_{ab}E_{ab}
}.
\label{eq:qwk}
\end{equation}

Errors spanning several ESI levels therefore receive greater weight than
adjacent-level errors.

\paragraph{Under-triage rate.}
Because larger ESI values indicate lower urgency, an encounter is under-triaged
when

\begin{equation}
\hat{y}_i>y_i.
\end{equation}

The model-specific under-triage rate reported in
Table~\ref{tab:dialogue-esi-and-closing} is

\begin{equation}
\operatorname{Under}
=
\frac{1}{N}
\sum_{i=1}^{N}
\mathbf{1}\!\left[\hat{y}_i>y_i\right].
\label{eq:undertriage}
\end{equation}

An ESI-5 encounter has no less urgent category and therefore cannot satisfy
this definition.

\subsection{Patient-Facing Closing Generation}
\label{app:closing-metrics}

Each generated closing is evaluated against the completed dialogue and the
held-out reference closing. The grounding metrics operate on atomic clinical
information units. Reference-similarity metrics operate on the generated and
reference texts, and Distinct-2 is computed over the complete set of generated
closings.

\paragraph{Atomic information units.}
Let \(K_i=\{k_{i,1},\ldots,k_{i,n_i}\}\) be the set of key clinical information
units associated with the reference closing for encounter \(i\). Each key unit
must be supported by the completed dialogue. Let
\(A_i=\{a_{i,1},\ldots,a_{i,m_i}\}\) be the atomic clinical claims extracted
from the generated closing.

The fixed evaluator determines whether a generated claim expresses a key unit
and whether it is supported by the dialogue. The same evaluator, matching
criteria, and prompts are applied to every model output.

\paragraph{Key Recall.}
A reference key unit is recovered when at least one generated claim expresses
the same clinical information. Encounter-level Key Recall is

\begin{equation}
\operatorname{KeyRecall}_i
=
\frac{
\sum_{k\in K_i}
\mathbf{1}\!\left[
\exists a\in A_i:
\operatorname{Match}(a,k)=1
\right]
}{
|K_i|
}.
\label{eq:key-recall-instance}
\end{equation}

The corpus score is the encounter-level macro-average:

\begin{equation}
\operatorname{KeyRecall}
=
\frac{1}{N}
\sum_{i=1}^{N}
\operatorname{KeyRecall}_i.
\label{eq:key-recall}
\end{equation}

This metric measures the coverage of clinically important information in the
patient-facing closing. Repeated statements of the same key unit receive one
credit.

\paragraph{Grounded Precision.}
A generated atomic claim is grounded when it is directly supported by the
completed dialogue. Encounter-level Grounded Precision is

\begin{equation}
\operatorname{GroundedPrecision}_i
=
\frac{
\sum_{a\in A_i}
\mathbf{1}\!\left[
\operatorname{Supported}(a,\widetilde H_i)=1
\right]
}{
|A_i|
},
\label{eq:grounded-precision-instance}
\end{equation}

where \(\widetilde H_i\) is the speaker-labeled dialogue provided to the model.
The corpus score is

\begin{equation}
\operatorname{GroundedPrecision}
=
\frac{1}{N}
\sum_{i=1}^{N}
\operatorname{GroundedPrecision}_i.
\label{eq:grounded-precision}
\end{equation}

Unsupported diagnoses, symptoms, test results, treatments, or other clinical
statements reduce precision. Administrative phrasing, greetings, and general
reassurance without a clinical assertion are not treated as atomic clinical
claims. A response containing no clinical claim receives a Grounded Precision
of zero.

\paragraph{BERTScore F1.}
BERTScore aligns contextualized token representations between the generated
closing \(g_i\) and reference closing \(r_i\). It calculates token-level
precision and recall using maximum cosine similarity and combines them into an
F1 score. We report the mean encounter-level BERTScore F1:

\begin{equation}
\operatorname{BERTScore}
=
\frac{1}{N}
\sum_{i=1}^{N}
\operatorname{BERTScoreF1}(g_i,r_i).
\label{eq:bertscore}
\end{equation}

\paragraph{ROUGE-L.}
ROUGE-L is based on the length of the longest common subsequence between the
generated and reference closings. We report the F-measure obtained from
longest-common-subsequence precision and recall, macro-averaged across
encounters:

\begin{equation}
\operatorname{ROUGE\text{-}L}
=
\frac{1}{N}
\sum_{i=1}^{N}
\operatorname{ROUGE\text{-}L}_{F}(g_i,r_i).
\label{eq:rouge-l}
\end{equation}

\paragraph{BLEU-4.}
BLEU-4 measures modified \(n\)-gram precision for \(n\in\{1,2,3,4\}\), together
with a brevity penalty:

\begin{equation}
\operatorname{BLEU\text{-}4}
=
\operatorname{BP}
\exp\left(
\frac{1}{4}
\sum_{n=1}^{4}\log p_n
\right),
\label{eq:bleu4}
\end{equation}

where \(p_n\) is the clipped precision of generated \(n\)-grams relative to the
references and \(\operatorname{BP}\) is the standard brevity penalty. BLEU-4 is
computed at the corpus level using one reference closing per encounter.

\paragraph{Distinct-2.}
Distinct-2 measures lexical diversity using the proportion of unique generated
bigrams in the full output corpus. Let \(\mathcal{G}_2\) be the multiset of all
bigrams in the generated closings. Then

\begin{equation}
\operatorname{Distinct\text{-}2}
=
\frac{
|\operatorname{Unique}(\mathcal{G}_2)|
}{
|\mathcal{G}_2|
}.
\label{eq:distinct2}
\end{equation}

Higher values indicate a larger proportion of unique bigrams across the
generated responses.

\subsection{Safety and Confidence Metrics}
\label{app:safety-metrics}

The safety analysis uses the ESI predictions and score vectors produced in
Experiment~II. Let \(\mathcal{M}\) denote the retained set of models with valid,
non-degenerate ESI predictions and confidence outputs.

\paragraph{ESI-specific under-triage.}
For recorded ESI level \(k\), the class-conditional under-triage rate for model
\(m\) is

\begin{equation}
\operatorname{Under}_{m,k}
=
\frac{
\sum_{i:y_i=k}
\mathbf{1}\!\left[\hat{y}_{i,m}>k\right]
}{
\sum_i \mathbf{1}\!\left[y_i=k\right]
}.
\label{eq:class-undertriage}
\end{equation}

\paragraph{Majority and unanimous under-triage.}
An encounter is majority under-triaged when more than half of the retained
models assign a less urgent ESI level:

\begin{equation}
\operatorname{MajorityUnder}_i
=
\mathbf{1}\!\left[
\sum_{m\in\mathcal{M}}
\mathbf{1}\!\left[\hat{y}_{i,m}>y_i\right]
>
\frac{|\mathcal{M}|}{2}
\right].
\label{eq:majority-under-instance}
\end{equation}

For recorded level \(k\), the majority under-triage rate is

\begin{equation}
\operatorname{MajorityUnder}_k
=
\frac{
\sum_{i:y_i=k}\operatorname{MajorityUnder}_i
}{
\sum_i\mathbf{1}\!\left[y_i=k\right]
}.
\label{eq:majority-under}
\end{equation}

An encounter is unanimously under-triaged when every retained model assigns a
less urgent level:

\begin{equation}
\operatorname{UnanimousUnder}_i
=
\mathbf{1}\!\left[
\sum_{m\in\mathcal{M}}
\mathbf{1}\!\left[\hat{y}_{i,m}>y_i\right]
=
|\mathcal{M}|
\right].
\label{eq:unanimous-under}
\end{equation}

\paragraph{Catastrophic under-triage.}
Catastrophic under-triage is defined as assigning ESI 4 or 5 to an encounter
recorded as ESI 1 or 2:

\begin{equation}
C_{i,m}
=
\mathbf{1}\!\left[
y_i\in\{1,2\}
\land
\hat{y}_{i,m}\in\{4,5\}
\right].
\label{eq:catastrophic-under}
\end{equation}

This definition is used for the confidence comparison in
Figure~\ref{fig:safety-undertriage}(b).

\paragraph{Reported confidence.}
Model-reported confidence is the maximum value in the parsed five-class score
vector:

\begin{equation}
c_{i,m}
=
\max_{k\in\{1,\ldots,5\}}p_{i,m,k}.
\label{eq:max-confidence}
\end{equation}

For model \(m\), mean confidence on correct predictions is

\begin{equation}
\overline{c}_{m}^{\,\mathrm{correct}}
=
\frac{
\sum_i
c_{i,m}\,
\mathbf{1}\!\left[\hat{y}_{i,m}=y_i\right]
}{
\sum_i
\mathbf{1}\!\left[\hat{y}_{i,m}=y_i\right]
},
\label{eq:correct-confidence}
\end{equation}

and mean confidence on catastrophic under-triage is

\begin{equation}
\overline{c}_{m}^{\,\mathrm{cat}}
=
\frac{
\sum_i c_{i,m}C_{i,m}
}{
\sum_i C_{i,m}
}.
\label{eq:catastrophic-confidence}
\end{equation}

The model-level sign test compares
\(\overline{c}_{m}^{\,\mathrm{cat}}\) and
\(\overline{c}_{m}^{\,\mathrm{correct}}\) across retained models. Models without
a catastrophic prediction do not contribute to this paired comparison.

\paragraph{Confidence-based correctness detection.}
For each model, we also evaluate whether maximum class confidence separates
correct from incorrect predictions. Prediction correctness is represented by

\begin{equation}
u_{i,m}
=
\mathbf{1}\!\left[\hat{y}_{i,m}=y_i\right].
\end{equation}

We calculate AUROC using \(c_{i,m}\) as the score and \(u_{i,m}\) as the binary
outcome. An AUROC of \(0.5\) corresponds to random ranking, while a value above
\(0.5\) indicates that correct predictions generally receive higher confidence
than incorrect predictions.

\end{document}